\documentclass{article}
\usepackage[preprint]{neurips_2026}
\usepackage[utf8]{inputenc} 
\usepackage[T1]{fontenc}    
\usepackage{url}            
\usepackage{booktabs}       
\usepackage{amsfonts}       
\usepackage{nicefrac}       
\usepackage{microtype}      
\usepackage{xcolor}         
\usepackage{graphicx}
\usepackage{subcaption}
\usepackage{tabularx}
\usepackage{amsmath}
\usepackage{amssymb}
\usepackage{mathtools}
\usepackage{amsthm}
\usepackage{multirow}
\usepackage{enumitem}
\usepackage{wrapfig}
\usepackage[table]{xcolor}
\definecolor{lightgray}{gray}{0.9}
\definecolor{lightblue}{RGB}{220,235,255}
\definecolor{citecolor}{HTML}{2980b9}
\definecolor{linkcolor}{HTML}{c0392b}
\usepackage[colorlinks, citecolor=citecolor, linkcolor=linkcolor]{hyperref}
\setcitestyle{open={(}, close={)}, aysep={, }}

\title{OnlineCache: Learning Dynamic Caching Policies with Error Correction for Efficient Diffusion Inference}

\author{%
  Zhikang Xie$^{1}$\thanks{Equal contribution.},
  Xichen Ye$^{1}$\footnotemark[1],
  Yifan Wu$^{1}$,
  Haoshen Yu$^{1}$,
  Li Chenan$^{1}$, \\
  \textbf{Peizhu Gong}$^{1}$,
  \textbf{Weizhong Zhang}$^{2}$,
  \textbf{Cheng Jin}$^{1,3}$\thanks{Corresponding author.}\\
  $^{1}$College of Computer Science and Artificial Intelligence, Fudan University\\
  $^{2}$School of Data Science, Fudan University\\
  $^{3}$Shanghai Key Laboratory of Intelligent Information Processing
}

\begin{document}
\maketitle
\begin{abstract}
Diffusion models have revolutionized generative tasks but incur high latency due to iterative denoising. While cache-based strategies accelerate inference by reusing intermediate features, they largely rely on static, sample-agnostic schedules. We argue that this rigidity overlooks two facts empirically validated in this paper: (i) generation difficulty varies across prompts, requiring adaptive resource allocation—complex inputs demand more computation while simpler ones require less; (ii) error sensitivity fluctuates across timesteps, where static policies may cache high-error steps or waste computation on low-error ones. We therefore propose \textbf{OnlineCache}, a dynamic caching framework that jointly learns \textit{when to cache} and \textit{how to correct} approximation errors. We leverage policy gradient to train a lightweight network for adaptive speed-quality trade-offs, and incorporate a learnable corrector to mitigate caching-induced errors. Both modules are jointly optimized under a bilevel optimization framework, with the policy targeting global generation quality and the corrector minimizing local errors. Our method automatically allocates computational resources across both samples and timesteps, improving overall generation quality. Extensive experiments demonstrate clear superiority. On FLUX.1-dev model, OnlineCache achieves nearly 3$\times$ speedup while preserving generation fidelity. On DiT and CogVideoX, it similarly delivers competitive acceleration without compromising quality; across all scenarios, it consistently outperforms existing cache-based acceleration baselines. 
\end{abstract}

\section{Introduction}
\label{introduction}
In recent years, diffusion models~\citep{DBLP:conf/icml/Sohl-DicksteinW15, DBLP:conf/nips/SongE19, DBLP:conf/nips/HoJA20, DBLP:conf/nips/DhariwalN21} have emerged as a dominant paradigm in generative AI, delivering unparalleled synthesis quality across a wide range of modalities, including images~\citep{DBLP:conf/cvpr/RombachBLEO22}, videos~\citep{DBLP:journals/corr/abs-2311-15127}, audio~\citep{DBLP:conf/iclr/KongPHZC21}, and 3D content~\citep{DBLP:conf/iclr/PooleJBM23}. This success is largely attributed to the adoption of highly scalable transformer-based backbones~\citep{DBLP:conf/iccv/PeeblesX23}. However, such performance gains come at the cost of substantial computational overhead. As these models become increasingly complex and large-scale, they exhibit slow inference and high deployment costs, posing serious challenges for real-time applications. 

To mitigate these computational bottlenecks, various acceleration techniques have been explored in recent years. Most existing approaches rely on modifying the underlying model, such as network pruning~\citep{DBLP:conf/nips/FangMW23}, low-bit quantization~\citep{DBLP:conf/cvpr/ShangYXW023}, and knowledge distillation~\citep{DBLP:journals/corr/abs-2101-02388}, all of which reduce inference cost by altering model structure or parameter representation. 
In contrast, an emerging line of work focuses on \textbf{cache-based acceleration}, which exploits the temporal redundancy across diffusion steps by reusing intermediate representations~\citep{DBLP:conf/cvpr/MaFW24, DBLP:conf/nips/MaFMW24, DBLP:journals/corr/abs-2406-01125}. Owing to its model-agnostic nature and minimal architectural intervention, cache-based methods offer a complementary and increasingly attractive alternative.

\begin{figure*}[t]
    \centering
    \begin{subfigure}[b]{0.495\textwidth}
        \centering
        \includegraphics[width=\linewidth]{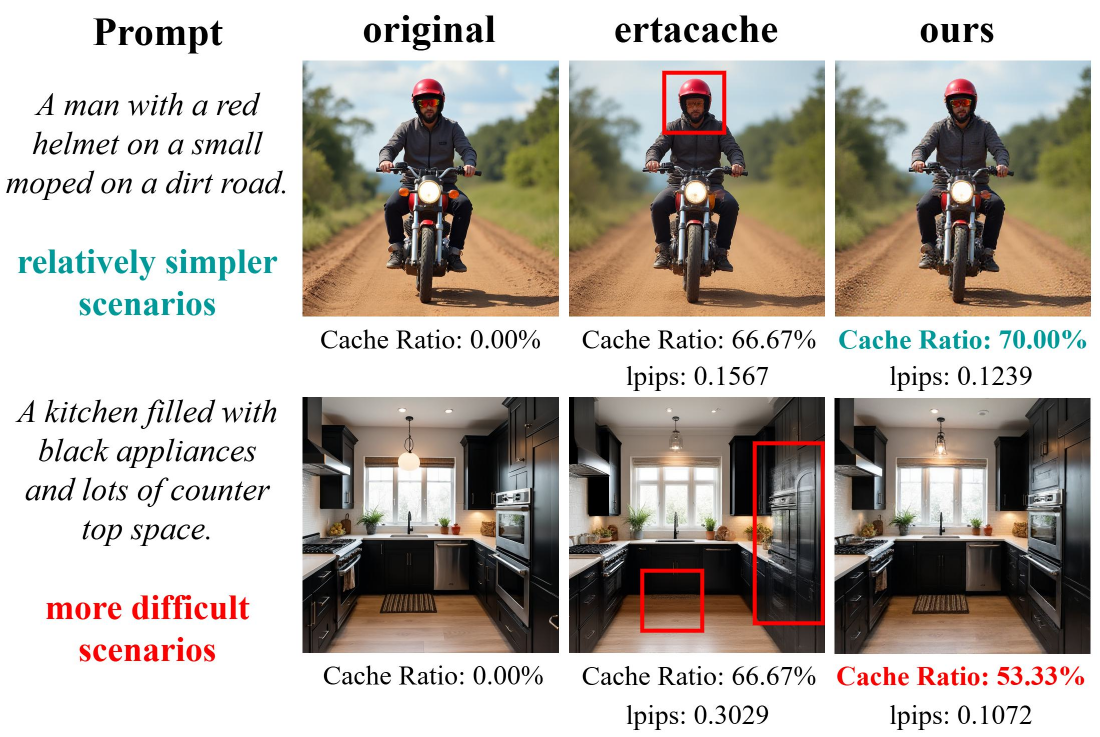}
        \caption{\textbf{Sample-level heterogeneity.} OnlineCache dynamically allocates computation across prompts of varying difficulty.}
        \label{fig:motivation_adaptivity}
    \end{subfigure}
    \hfill
    \begin{subfigure}[b]{0.49\textwidth}
        \centering
        \includegraphics[width=\linewidth]{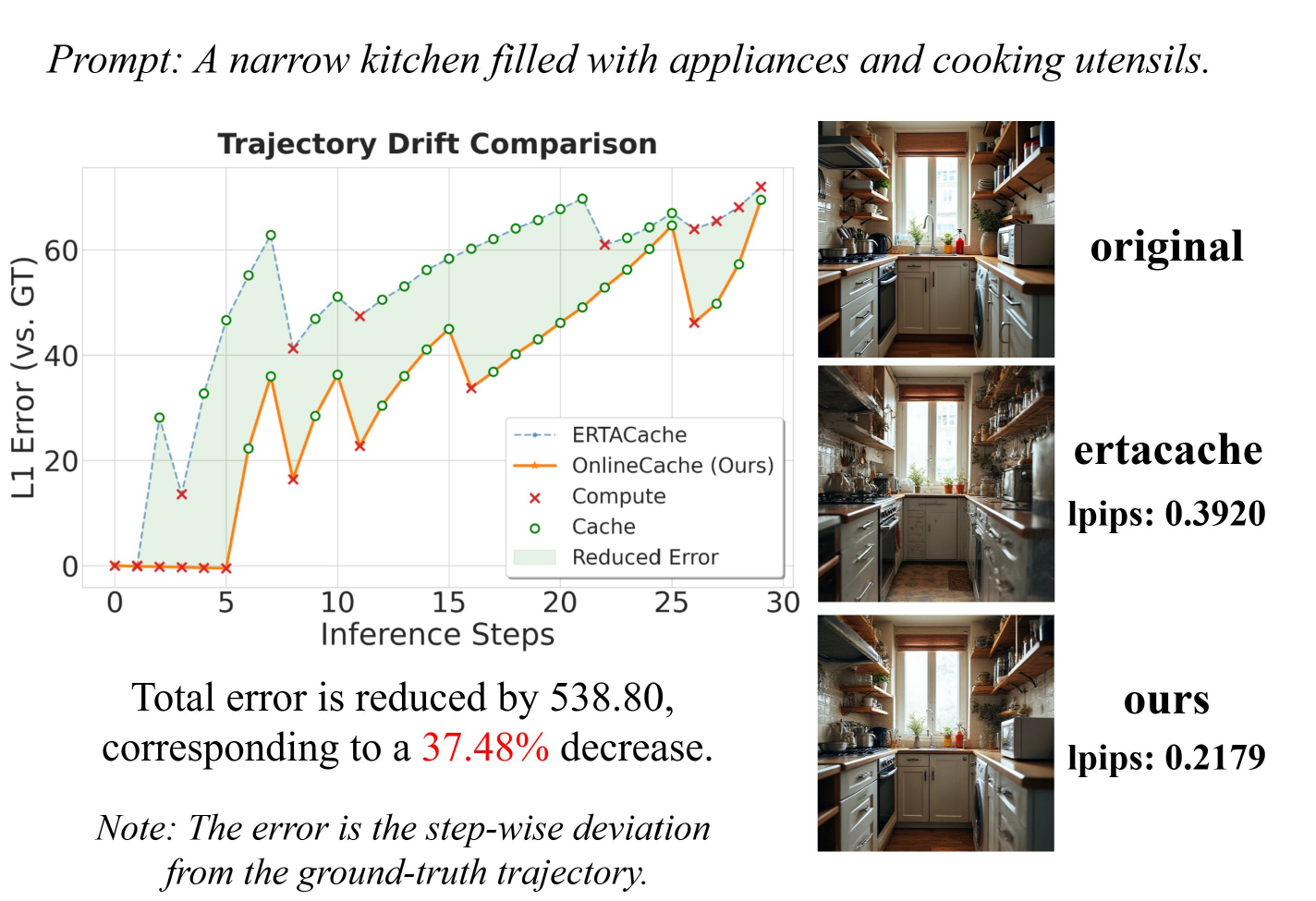}
        \caption{\textbf{Timestep-level heterogeneity.} OnlineCache adaptively identifies timesteps that are suitable for caching.}
        \label{fig:motivation_fixed_ratio}
    \end{subfigure}
    \caption{Comparison of static strategy (ERTACache) and ours (OnlineCache) in different senarios of diffusion acceleration.
    \textbf{(a)} Under different prompt complexities, ERTACache adopts a uniform caching strategy, whereas OnlineCache dynamically adapts (e.g., 70.00\% vs. 53.33\% cache ratio), achieving improved generation quality under a comparable average speedup. 
    \textbf{(b)} Under the same 66.67\% caching ratio, ERTACache's "one-policy-fits-all" strategy is suboptimal. In contrast, OnlineCache identifies critical timesteps and achieves a 37.48\% reduction in $L_1$ error, along with substantially improved perceptual quality.}
    \label{fig:motivation}
\end{figure*}

However, most existing cache-based methods rely on static or heuristic-driven rules to determine intermediate feature reuse, which fundamentally constrains their flexibility and performance. Static strategies such as ERTACache~\citep{DBLP:journals/corr/abs-2508-21091} adopt a fixed caching schedule—applying the same set of cached steps to all samples regardless of input complexity. More adaptive methods like TeaCache~\citep{DBLP:conf/cvpr/Liu0W0QZZY025}, which utilize heuristic threshold, still remain suboptimal due to their greedy, local decisions that lack global quality awareness. 
Through an empirical analysis of diffusion trajectories (as evidenced in Figure~\ref{fig:motivation}), we argue that such rigid, rule-based policies are insufficient to handle the intrinsic heterogeneity of the denoising process, which manifests at two levels: 
\begin{itemize}[label=\(\triangleright\), leftmargin=*, itemsep=2pt]
    \item \textbf{Sample-level heterogeneity:} 
    Figure~\ref{fig:motivation_adaptivity} demonstrates generation difficulty varies significantly across prompts. While static method allocates computation uniformly, our dynamic policy adaptively allocates resources across samples, assigning more computation to difficult cases and less to easy ones, thereby achieving better performance under a comparable average speedup. 
    \item \textbf{Timestep-level heterogeneity:} Figure~\ref{fig:motivation_fixed_ratio} shows that error sensitivity fluctuates along the denoising trajectory. Static method inevitably caches sensitive steps while wasting computation on redundant ones, resulting in suboptimal quality. In contrast, OnlineCache identifies critical timesteps and significantly reduces overall errors (37.48\% $L_1$ reduction; LPIPS improves from 0.3920 to 0.2179).
\end{itemize}
These observations reveal a fundamental mismatch between rigid scheduling and complex denoising. We argue that effective acceleration should move beyond fixed rules and adopt a dynamic, instance-aware policy. \textbf{To this end, we propose OnlineCache}, a framework that formulates the caching schedule as a sequential decision-making problem. Instead of relying on heuristics, we employ a lightweight policy network to evaluate latent states at each denoising timestep and decide whether to reuse cached representations or perform full computation. The policy is optimized via policy gradient~\citep{DBLP:conf/nips/SuttonMSM99} to maximize a trajectory-level reward, endowing the model with a global perspective for efficient allocation of computational budgets. 
Specifically, we introduce two variants of OnlineCache:
(i) a \textbf{vanilla} variant that trains a standalone policy network to learn adaptive caching decisions, and
(ii) a \textbf{bilevel optimization} (BLO) variant that jointly trains the policy while optimizing an error corrector to compensate for caching-induced trajectory deviations. 

In the BLO variant, policy optimization is formulated as the outer objective to effectively balance the trade-off between acceleration and generation quality, while the error corrector is optimized as the inner objective to minimize local error. This bilevel coupling ensures that the policy learns to skip computation strictly conditional on the corrector’s capacity to repair the induced error, leading to a superior Pareto frontier between efficiency and quality. Moreover, both the policy and the corrector are designed as lightweight MLPs, introducing negligible inference-time latency in practice.

\begin{figure}[t]
    \centering
    \includegraphics[width=\linewidth]{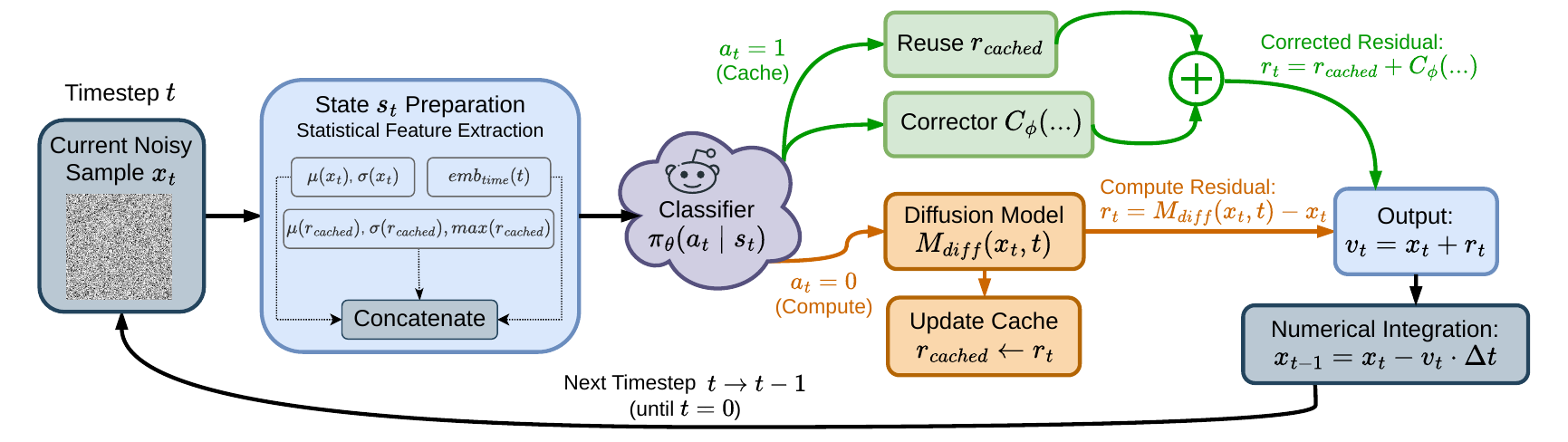}
    \caption{\textbf{Overall inference pipeline of OnlineCache}, where the policy network dynamically determines caching decisions and the corrector compensates for cache-induced approximation errors.}
    \label{fig:inference_flowchart}
\end{figure}

Our core contributions can be summarized as follows:
\begin{itemize}[label=\(\triangleright\), leftmargin=*, itemsep=2pt]
    \item We propose \textbf{OnlineCache}, a dynamic, instance-aware cache-based acceleration framework, with two variants: a policy-only vanilla model and a bilevel optimization model that jointly trains policy with an error corrector, enabling a globally effective speed-quality trade-off. 
    \item We develop a practical training pipeline for OnlineCache, incorporating multiple optimization strategies and engineering techniques to ensure robust convergence, stable joint training, and efficient \textbf{plug-and-play} inference, without any architectural modifications to the backbone. Code and trained policy networks for several mainstream diffusion models will be released.
    \item We conduct extensive experiments and ablation studies to validate the \textbf{effectiveness} of our approach. OnlineCache accelerates 1.88$\times$ on \textsc{DiT-XL/2} while maintaining competitive FID relative to full-step inference. On \textsc{FLUX.1-dev}, it attains nearly 3$\times$ speedup with consistently high generation quality, surpassing prior cache-based methods. It also extends seamlessly to video generation.
\end{itemize}

\section{Related Works}
\paragraph{Cache-based Acceleration.}
Cache-based acceleration leverages temporal redundancy across diffusion steps to reduce inference cost without modifying the backbone model. Early methods mostly adopt fixed caching schedules or architecture-specific reuse, such as periodic feature reuse in DeepCache~\citep{DBLP:conf/cvpr/MaFW24} and FORA~\citep{DBLP:journals/corr/abs-2407-01425}. To improve robustness, later methods introduce calibration or correction mechanisms, including Block Caching~\citep{DBLP:conf/cvpr/WimbauerWSDHHSZ24}, $\Delta$-DiT~\citep{DBLP:journals/corr/abs-2406-01125}, and ERTACache~\citep{DBLP:journals/corr/abs-2508-21091}. More recent work moves toward dynamic and learning-based policies, exemplified by TeaCache~\citep{DBLP:conf/cvpr/Liu0W0QZZY025}, DiCache~\citep{DBLP:journals/corr/abs-2508-17356}, and router-based or compensation-based approaches~\citep{DBLP:conf/nips/MaFMW24, DBLP:journals/corr/abs-2505-20353}. Our method, OnlineCache, also belongs to this learning-based cache-acceleration family.

Beyond caching, diffusion acceleration has also been explored through pruning, quantization, distillation, and improved samplers. We refer readers to Appendix~\ref{appendix_related_works} for a more comprehensive discussion.

\section{Methodology}
In this section, we detail \textbf{OnlineCache}, a dynamic caching framework for diffusion acceleration. We formulate the caching decision as a sequential decision-making problem and solve it via Policy Gradient~\citep{DBLP:conf/nips/SuttonMSM99}.
Specifically, we introduce two versions: (i) a \textbf{vanilla} variant trains a separate policy network and (ii) a \textbf{bilevel optimization} variant that jointly trains the policy while tuning an error corrector for caching-induced deviations. 

\subsection{Problem Formulation}
We consider a diffusion-based generative process that progressively denoises a latent variable from $x_N$ to $x_0$ over a sequence of timesteps $\{t_N, \dots, t_1\}$. During inference, a diffusion model $\mathcal{M}_\text{diff}$ takes the current noisy sample $x_t$ at timestep $t$ and predicts the next-step intermediate state. Here, we take the $v$-prediction parameterization as an example, while the $x$- and $\epsilon$-prediction variants follow analogously. Specifically, the model output can be formulated in a \textit{residual} form as: 
\begin{equation}
    v_t = \mathcal{M}_\text{diff}(x_t, t) = x_t + r_t,
\end{equation}
where $r_t = v_t - x_t$ denotes the residual at timestep $t$. The sample at the next timestep $x_{t-1}$ is then obtained by applying a numerical integration step: $x_{t-1} = x_t - v_t \cdot \Delta t$, where $\Delta t$ denotes the integration step size. Previous works, such as ERTACache~\citep{DBLP:journals/corr/abs-2508-21091}, have consistently shown that in diffusion process, the residual feature $r_t$ often evolves smoothly across adjacent timesteps. This temporal redundancy suggests that $r_t$ can be well approximated by a previously computed and cached residual from a recent timestep, denoted as $r_{\text{cached}}$.

We introduce a binary variable $a_t \in \{0, 1\}$ per timestep: $a_t=1$ signifies \textit{caching} (skip the heavy computation and reuse $r_{\text{cached}}$), and $a_t=0$ signifies \textit{computing}. The update rule is defined as:
\begin{equation}
\label{eq:vt-cal}
\begin{aligned}
    v_t &= x_t + \hat{r}_t, \quad
    \mbox{where }\hat{r}_t =
    \begin{cases}
        r_{\text{cached}}, & \text{if } a_t = 1, \\
        \mathcal{M}_\text{diff}(x_t, t) - x_t, & \text{if } a_t = 0.
    \end{cases}
\end{aligned}
\end{equation}

When $a_t=0$, we update the cache with the exact residual: $\hat{r}_t \rightarrow r_{\text{cached}}$. This mechanism allows us to skip the expensive computation of $\mathcal{M}_\text{diff}$ entirely when the cached residual is deemed sufficient.

\subsection{Policy Network Architecture and Design}
\label{rl-based-decision-policy}
\begin{wrapfigure}{r}{0.34\columnwidth}
    \vspace{-0.2in}
    \centering
    \includegraphics[width=\linewidth]{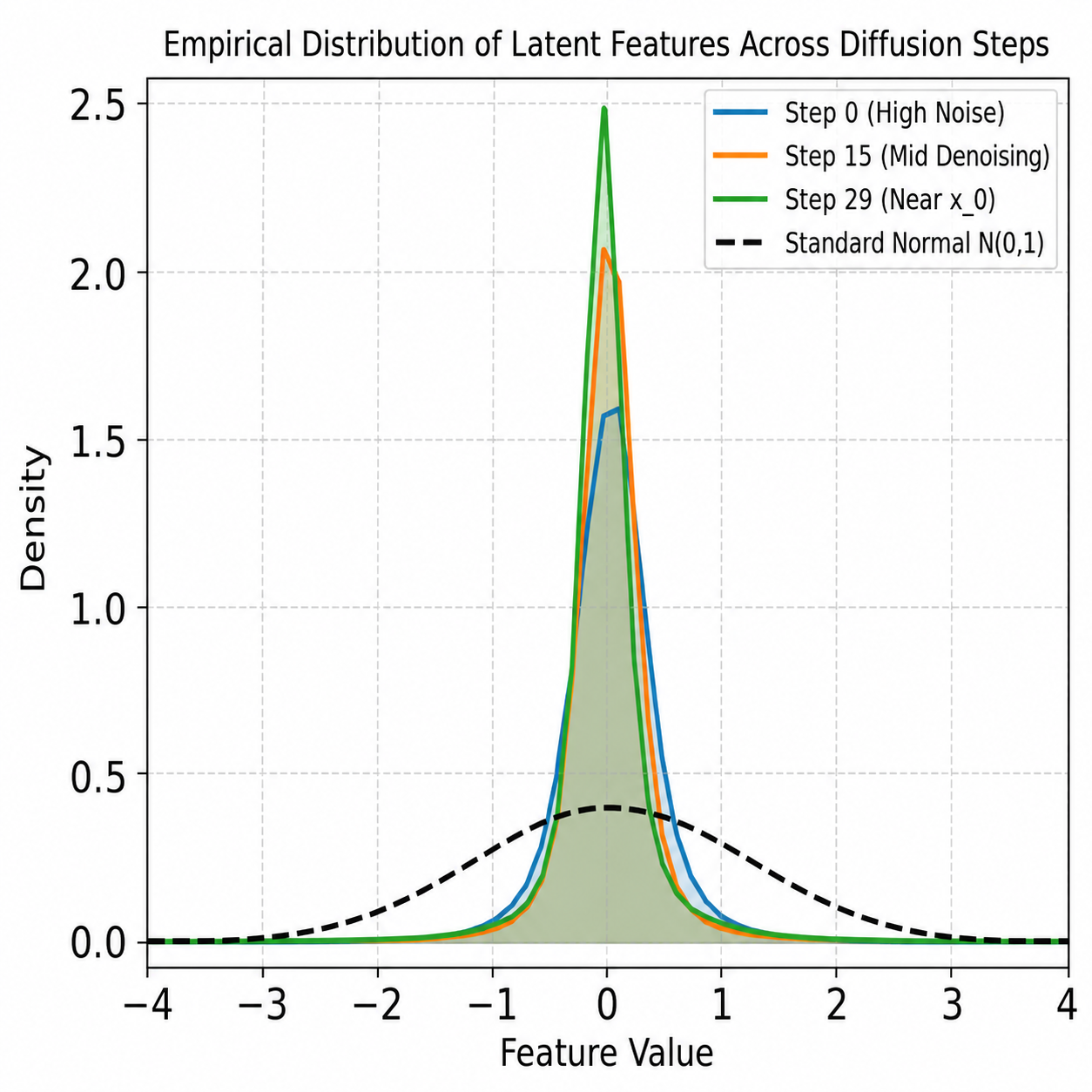}
    \vspace{-0.15in}
    \caption{Statistical feature design for policy input. We extract $(\mu(\cdot), \sigma(\cdot))$ to characterize the latent distribution, motivated by the approximately Gaussian behavior of latent representations in diffusion models.}
    \label{fig:gaussian_feature}
    \vspace{-0.2in}
\end{wrapfigure}

The decision to reuse residuals is non-trivial: aggressive caching may cause error accumulation and quality degradation, while conservative reuse diminishes efficiency gains. To balance these factors, we formulate caching as a Markov Decision Process (MDP) and train a lightweight policy $\pi_\theta$ to maximize a reward signal balancing quality and speedup. Crucially, the effectiveness of the policy depends on a well-designed state representation input, guided by the following two key observations: 
\begin{itemize}[label=\(\triangleright\), leftmargin=*, itemsep=2pt]
    \item \textbf{Statistical Sufficiency.} Latent features during diffusion often exhibit Gaussian-like behavior, making the channel-wise mean $\mu(\cdot)$ and standard deviation $\sigma(\cdot)$ informative global summaries. This is \textbf{theoretically} grounded in Latent Diffusion Models, where the VAE imposes a KL constraint that regularizes the latent space toward $\mathcal{N}(0, I)$; combined with the Gaussian initialization at $x_T$, intermediate states naturally retain Gaussian-like properties. \textbf{Empirically}, as shown in Figure~\ref{fig:gaussian_feature}, analysis of hidden states from the \textsc{FLUX.1-dev}~\citep{DBLP:journals/corr/abs-2506-15742} model across early, middle, and final timesteps confirms this behavior. Accordingly, we extract $\mu(\cdot)$ and $\sigma(\cdot)$ from both the current latent $x_t$ and the cached residual $r_{\text{cached}}$. In addition, we include $\max(|r_{\text{cached}}|)$ to capture salient localized updates that may be obscured by average statistics.
    \item \textbf{Temporal Awareness.} Diffusion dynamics are non-stationary, evolving from coarse structural formation to fine-grained refinement. The time embedding $\text{emb}_{time}(t)$ supplies phase information, enabling the policy to adapt its caching behavior across different denoising stages.
\end{itemize}

Based on the above analysis, let $D$ be the channel dimension of the latent space. The state $s_t$ provided to the policy network $\pi_\theta(a_t | s_t)$ is a concatenated vector in $\mathbb{R}^{6 \times D}$:
\begin{equation}
    s_t = \text{Concat}\left( 
    \begin{aligned}
    &\mu(x_{t}), \sigma(x_{t}), \\
    &\mu(r_{\text{cached}}), \sigma(r_{\text{cached}}), \max(|r_{\text{cached}}|), \\
    &\text{emb}_{time}(t)
    \end{aligned}
    \right).
\end{equation}

The policy $\pi_\theta$ is parameterized by a lightweight Multilayer Perceptron(MLP). The architecture consists of:
(i) An input projection layer compressing the feature dimension from $6 \times D$ to $2 \times N_{\text{hidden}}$, followed by LayerNorm and ReLU.
(ii) A bottleneck layer reducing dimensions to $N_{\text{hidden}}/2$, followed by ReLU.
(iii) A final projection to a scalar probability $p_t$ via a Sigmoid activation. The action is then sampled stochastically with $a_t \sim \text{Bernoulli}(p_t)$. To assess the validity of these design choices, we perform a controlled ablation over the state components, as analyzed in Section~\ref{ablation:input_state_decision}.

\subsection{Reward Design and Policy Optimization}
\label{reward-engineering}
We design a composite reward function $R(\tau)$ evaluated over the full generation trajectory $\tau$. The reward encourages high visual fidelity while enforcing a specific acceleration rate. 
\begin{itemize}[label=\(\triangleright\), leftmargin=*, itemsep=2pt]
    \item \textbf{Quality Reward ($R_{\text{q}}$)}: We measure the perceptual similarity between the image generated by our dynamic policy, $I_{\text{pred}}$, and the ground-truth image generated with full compute, $I_{\text{gt}}$. We utilize the combination of LPIPS~\citep{DBLP:conf/cvpr/ZhangIESW18} and SSIM~\citep{DBLP:journals/tip/WangBSS04}:
    \begin{equation}
        R_{\text{q}} = \lambda_q \left( 1 - \mathcal{L}_{\text{LPIPS}}(I_{\text{gt}}, I_{\text{pred}}) + \mathcal{L}_{\text{SSIM}}(I_{\text{gt}}, I_{\text{pred}}) \right).
    \end{equation}
    \item \textbf{Acceleration Constraint ($R_{\text{a}}$)}: To enforce a target compute budget, we penalize deviations from a target cache ratio $\rho_{\text{target}}$ (e.g., $0.5$, aiming for the policy to converge to caching roughly 50\%):
    \begin{equation}
    \label{eq:acceleration_constraint}
        R_{\text{a}} = -\lambda_a \left| \frac{1}{T} \sum_{t=1}^T a_t - \rho_{\text{target}} \right|.
    \end{equation}
    \item \textbf{Stability Penalty ($R_{\text{s}}$)}: To prevent error explosion from consecutive caching, we apply a penalty if the number of consecutive skips $k_t$ exceeds a tolerance threshold $k_{\text{tol}}$:
    \begin{equation}
        R_{\text{s}} = - \lambda_s \left( \sum_{t=1}^T \mathbf{1}_{(k_t > k_{\text{tol}})} \cdot (k_t - k_{\text{tol}}) \right).
    \end{equation}
\end{itemize}

The total reward for trajectory $\tau$ is the sum: $R(\tau) = R_{\text{q}} + R_{\text{a}} + R_{\text{s}}$. Since $R_{\text{a}}$ and $R_{\text{s}}$ are always negative, they serve as penalty terms. We optimize the policy parameters $\theta$ using the REINFORCE algorithm (Monte Carlo Policy Gradient). To bridge the gap between raw rewards and stable gradient updates, we transform the total reward into a normalized advantage score and incorporate entropy regularization. For a detailed derivation of the reverse gradient flow, please refer to Appendix~\ref{appendix1_gradient_derivation}. 

However, the raw reward signals $R(\tau)$ can vary significantly in magnitude, leading to high variance in gradient estimation. In practice, we follow a standard approach and compute a normalized advantage score for each trajectory relative to the current batch $\mathcal{B}$.
Let $R(\tau_i)$ be the cumulative reward for the $i$-th trajectory in a batch of size $B$. The advantage $A(\tau_i)$ for trajectory $i$ is then standardized:
\begin{equation}
\label{eq:advantage_norm}
    A(\tau_i) = \frac{R(\tau_i) - \frac{1}{B} \sum_{j=1}^B R(\tau_j)}{\sqrt{\frac{1}{B} \sum_{j=1}^B \left(R(\tau_j) - \frac{1}{B} \sum_{k=1}^B R(\tau_k)\right)^2 + \epsilon}},
\end{equation}
where $\epsilon$ is a small constant for numerical stability. This normalization assigns positive rewards to above-average trajectories and penalizes below-average ones.

To encourage exploration and prevent the policy from collapsing to a deterministic behavior early in training, we introduce an entropy bonus. 
Since our policy $\pi_\theta(s_t)$ outputs a scalar probability $p_t$ for the Bernoulli distribution, the entropy at timestep $t$ can be defined as:
\begin{equation}
\label{eq:entropy_term}
    \mathcal{H}(p_t) = - \left[ p_t \log p_t + (1 - p_t) \log (1 - p_t) \right].
\end{equation}

Finally, the total loss $\mathcal{L}(\theta)$ combines the policy gradient and entropy terms. We introduce $\lambda_e$ to control the exploration-exploitation trade-off, for a batch of trajectories, the objective is:
\begin{equation}
\label{eq:total-loss}
\begin{aligned}
    \mathcal{L}(\theta)
    = \frac{1}{B} \sum_{i=1}^B \Bigg[
        - \left( \sum_{t=1}^T \log \pi_\theta(a_{i,t} \mid s_{i,t}) \right) \cdot A(\tau_i) 
        - \lambda_e \frac{1}{T} \sum_{t=1}^T \mathcal{H}(\pi_\theta(s_{i,t}))
    \Bigg].
\end{aligned}
\end{equation}

\subsection{Bilevel Optimization Framework}
\label{bilevel-optimization-framework}
To further improve performance, we introduce an error corrector implemented as a lightweight MLP, which compensates for the drift induced by cache reuse. We formulate the joint training as a bilevel optimization problem. The framework consists of an outer-loop objective that optimizes the policy and a inner-loop objective that minimizes the local error. Formally, let $\theta$ and $\phi$ denote the parameters of the policy $\pi_\theta$ and corrector $\mathcal{C}_\phi$, respectively. The optimization problem can be defined as:
\begin{equation}
\begin{aligned}
\min_{\theta} \quad 
& \mathcal{L}_{\text{outer}}(\theta, \phi^*(\theta)) 
= - \mathbb{E}_{\tau \sim \pi_\theta} [R(\tau)] \quad\quad
\text{s.t.,} \quad\phi^*(\theta) \in \arg \min_{\phi} \mathcal{L}_{\text{inner}}(\theta, \phi),
\end{aligned}
\end{equation}

where $R(\tau)$ is the cumulative reward defined in Section~\ref{reward-engineering}.

\subsubsection{Outer-loop: Policy Optimization}
In the outer loop, we optimize the policy parameters $\theta$ while keeping the error corrector $\phi$ fixed, which represents the optimal solution $\phi^*$ under the current policy. The policy network learns to make binary caching decisions $a_t$ based on observed latent statistics. The optimization objective is to minimize the negative expected reward via the policy gradient defined in Equation~\ref{eq:total-loss}, where $A(\tau_i)$ denotes the normalized advantage of the trajectory generated through the interaction between the current policy and the corrector.
The update of $\theta$ can be expressed as:
\begin{equation}
\label{eq:implicit-gradient}
    \frac{d\mathcal{L}_{\text{outer}}}{d\theta} = \underbrace{\frac{\partial \mathcal{L}_{\text{outer}}}{\partial \theta}}_{\text{Direct Gradient}} + \underbrace{\frac{\partial \mathcal{L}_{\text{outer}}}{\partial \phi^*} \cdot \frac{d\phi^*(\theta)}{d\theta}}_{\text{Implicit Gradient}}.
\end{equation}
The \emph{direct gradient} term characterizes how variations in the policy $\theta$ directly influence the reward $R(\tau)$. 
In contrast, the \emph{implicit gradient} reflects the fact that changing $\theta$ induces a shift in the inner-level optimum $\phi^*$, since the corrector is trained conditional on the current caching decisions. This variation in $\phi^*$ in turn influences the outer-level reward.

\subsubsection{Inner-loop: Corrector Training}
The inner loop aims to find the optimal corrector $\phi^*$ that compensates for the numerical errors introduced when the policy network chooses to skip the transformer computation ($a_t=1$). When a step is cached, the accelerated output is calculated as $\hat{v}_t(\phi) = v_t + \mathcal{C}_\phi(x_t, r_{\text{cached}}, \text{emb}_{time}(t))$, where $v_t$ is computed according to Equation~\ref{eq:vt-cal} under the condition $a_t=1$. Given the pre-computed ground-truth intermediate states $v_t^{gt}$ (Appendix~\ref{appendix2_ablation_on_corrector_training_target} reports an ablation on the choice of corrector training target), the inner objective $\mathcal{L}_{\text{inner}}$ is defined as the Mean Squared Error over the cached steps: 
\begin{equation}
\label{eq:inner_loop}
\mathcal{L}_{\text{inner}}(\theta, \phi) = \sum_{t \,:\, a_t = 1}
\left\| \hat{v}_t(\phi) - v_t^{\mathrm{gt}} \right\|_2^2.
\end{equation}

\subsubsection{Practical BLO Implementation}
Computing the exact implicit gradient $\frac{d\phi^*(\theta)}{d\theta}$ in Equation~\ref{eq:implicit-gradient} is intractable in our setting for two reasons: (i) The dependency of the inner objective $\mathcal{L}_{\text{inner}}$ on the policy parameters $\theta$ is mediated by the discrete sampling of actions $a_t \sim \pi_\theta$. This non-differentiable sampling operation prevents the flow of gradients from the inner-loop loss back to the policy parameters via standard backpropagation. (ii) Even with continuous relaxation tricks (e.g., Gumbel-Softmax), calculating the exact implicit gradient typically involves computing Hessian-vector products, which imposes a prohibitive memory and computational burden given the high-dimensional context of diffusion models. 

Therefore, we adopt a \textbf{First-Order Approximation} of the BLO objective. Specifically, we simplify the optimization by alternating between the two levels: (i) Policy update (Outer-loop): We update $\theta$ using the policy gradient, assuming the current corrector $\phi$ is a localized constant approximation of the optimal $\phi^*(\theta)$. (ii) Corrector training (Inner-loop): We update $\phi$ via supervised fine-tuning based on the trajectories generated by the new policy $\pi_\theta$. This alternating scheme can be viewed as a coordinate descent approach to the BLO problem. By iteratively updating the policy and adapting the corrector to shifts in the action distribution, the framework converges to a joint equilibrium, enabling our OnlineCache to deliver substantial speedups while preserving generation quality.

\section{Experiments}

We design three sets of main experiments to evaluate our proposed OnlineCache: 
\begin{itemize}[label=\(\triangleright\), leftmargin=*, itemsep=2pt]
    \item \textsc{FLUX.1-dev} (Targeting SOTA Caching Methods \ref{ex:flux-ex}):  
    For image generation, we primarily compare against the SOTA caching methods and further provide a thorough generalization analysis.
    \item \textsc{DiT-XL/2} (Targeting Learning-Based Methods \ref{ex:dit-ex}):  
    OnlineCache introduces an extra training phase. We therefore compare with representative learning-based cache methods (FastCache, L2C), demonstrating that our approach achieves superior performance even within trainable frameworks.
    \item \textsc{CogVideoX-2b} (Targeting Cross-Modality Generalization \ref{ex:cogvideox-ex}):  
    For video generation, our goal is to validate the modality-agnostic nature of OnlineCache. Comparing with the strong baseline TeaCache is sufficient to show effective generalization beyond image diffusion models.
\end{itemize}

\subsection{\textsc{FLUX.1-dev} Experiment}
\label{ex:flux-ex}
\begin{table*}[h]
\centering
\small
\caption{\textbf{Quantitative comparison on FLUX}. We evaluate FLUX-series models on diverse datasets to demonstrate the strong performance and generalization capability of OnlineCache. Lat and CR denote Latency and Cache Ratio, respectively. Rows with the same background color denote comparable settings, and \textbf{bold} indicates the best result within each group.}
\label{tab:comparison-flux}
\begin{minipage}[t]{0.495\linewidth}
\resizebox{\columnwidth}{!}{
\begin{tabular}{lcccccc}
\hline
\addlinespace[0.1em]
\multirow{2}{*}{\textbf{Method}} & \multicolumn{2}{c}{\textbf{Efficiency}} & \multicolumn{3}{c}{\textbf{Visual Quality}} & \multirow{2}{*}{\textbf{CR}} \\
\cline{2-6}
\addlinespace[0.2em]
& \textbf{Speed}~$\uparrow$ & \textbf{Lat}~$\downarrow$ & \textbf{LPIPS}~$\downarrow$ & \textbf{SSIM}~$\uparrow$ & \textbf{PSNR}~$\uparrow$ & \\
\hline
\addlinespace[0.1em]
\multicolumn{7}{c}{\textit{\textbf{Block1: FLUX.1-dev, MSCOCO, 512$\times$512, 800 samples}}} \\[1pt]
FLUX(30 steps)  & 1.00$\times$ & 5.099 & --    & --    & --    & 0.00\% \\
\rowcolor{green!8} ERTACache   & 2.68$\times$ & 1.900 & 0.252 & \textbf{0.739} & 19.852 & 66.67\% \\
\rowcolor{green!8} TeaCache    & 2.60$\times$ & 1.960 & 0.329 & 0.669 & 17.006 & - \\
\rowcolor{green!8} OC-BLO(Ours)& \textbf{2.96$\times$} & 1.725 & \textbf{0.245} & \textbf{0.739} & \textbf{21.442} & 72.22\%\\
\rowcolor{lightblue} TeaCache    & 1.40$\times$ & 3.646 & 0.142 & 0.831 & 22.453 & - \\
\rowcolor{lightblue} OC-BLO(Ours) &\textbf{2.07$\times$} & 2.463 & \textbf{0.125} & \textbf{0.850} & \textbf{24.768} & 56.32\%\\
\hline
\addlinespace[0.1em]
\multicolumn{7}{c}{\textit{\textbf{Block2: FLUX.1-dev, MSCOCO, 1024$\times$1024, 500 samples}}} \\[1pt]
FLUX(30 steps) &1.00$\times$&15.470 & -- & -- & -- & 0.00\% \\
\rowcolor{green!8} FLUX(12 steps) &2.46$\times$&6.291&0.390&0.679&16.619& 60.00\% \\
\rowcolor{green!8} OC-BLO(Ours) &\textbf{3.25$\times$}&4.758&\textbf{0.374}&\textbf{0.694}&\textbf{20.265}&74.81\%\\
\rowcolor{lightblue} ERTACache &2.62$\times$ &5.912&0.281&0.755&20.382& 66.67\% \\
\rowcolor{lightblue} OC-BLO(Ours) &\textbf{2.75$\times$}&5.631&\textbf{0.275}&\textbf{0.762}&\textbf{21.343}&68.38\%\\
\hline
\addlinespace[0.1em]
\multicolumn{7}{c}{\textit{\textbf{Block3: FLUX.1-schnell, Parti-Prompts, 512$\times$512, 1632 samples}}} \\[1pt]
\rowcolor{green!8} FLUX(4 steps) &1.00$\times$&0.884 & -- & -- & -- & 0.00\% \\
\rowcolor{green!8} OC-BLO(Ours)  &1.31$\times$&0.674&0.070&0.844&23.261&34.40\%\\
\rowcolor{green!8} OC-BLO(Ours)  &1.66$\times$&0.534&0.131&0.671&22.649&57.72\%\\
\rowcolor{lightblue} FLUX(6 steps) &1.00$\times$&1.212 & -- & -- & -- & 0.00\% \\
\rowcolor{lightblue} OC-BLO(Ours)  &2.31$\times$&0.525&0.197&0.574&20.232&74.59\%\\
\hline
\end{tabular}
}
\end{minipage}\hfill
\begin{minipage}[t]{0.495\linewidth}
\resizebox{\columnwidth}{!}{
\begin{tabular}{lcccccc}
\hline
\addlinespace[0.1em]
\multirow{2}{*}{\textbf{Method}} & \multicolumn{2}{c}{\textbf{Efficiency}} & \multicolumn{3}{c}{\textbf{Visual Quality}} & \multirow{2}{*}{\textbf{CR}} \\
\cline{2-6}
\addlinespace[0.2em]
& \textbf{Speed}~$\uparrow$ & \textbf{Lat}~$\downarrow$ & \textbf{LPIPS}~$\downarrow$ & \textbf{SSIM}~$\uparrow$ & \textbf{PSNR}~$\uparrow$ & \\
\hline
\addlinespace[0.1em]
\multicolumn{7}{c}{\textit{\textbf{Block4: FLUX.1-dev, Parti-Prompts, 512$\times$512, 1632 samples}}} \\[1pt]
FLUX(30 steps) &1.00$\times$&5.090 & -- & -- & -- & 0.00\% \\
\rowcolor{green!8} FLUX(12 steps) &2.31$\times$&2.199&0.425&0.587&13.958&60.00\%\\
\rowcolor{green!8} TaylorSeer(N=6,O=1) &\textbf{3.18$\times$}&1.599&0.405&0.578&15.033&76.68\%\\
\rowcolor{green!8} TaylorSeer(N=6,O=2) &2.97$\times$&1.714&0.383&0.593&16.009&76.68\%\\
\rowcolor{green!8} TaylorSeer(N=5,O=1) &2.81$\times$&1.812&0.341&0.636&17.795&73.33\%\\
\rowcolor{green!8} TaylorSeer(N=5,O=2) &2.67$\times$&1.904&0.338&0.638&18.370&73.33\%\\
\rowcolor{green!8} OC-BLO(Ours) &\textbf{3.18$\times$}&1.601&\textbf{0.316}&\textbf{0.668}&\textbf{19.928}&73.39\%\\
\rowcolor{lightblue} ERTACache &2.68$\times$ &1.896&0.251&\textbf{0.746}&20.319& 66.67\% \\
\rowcolor{lightblue} TaylorSeer(N=4,O=1) &2.45$\times$&2.077&0.274&0.689&19.195&70.00\%\\
\rowcolor{lightblue} TaylorSeer(N=4,O=2) &2.38$\times$&2.142&0.267&0.693&19.690&70.00\%\\
\rowcolor{lightblue} OC-BLO(Ours) &\textbf{2.93$\times$}&1.739&\textbf{0.248}&0.740&\textbf{21.286}&70.92\%\\
\hline
\addlinespace[0.1em]
\multicolumn{7}{c}{\textit{\textbf{Block5: FLUX.1-dev, Parti-Prompts, 512$\times$512, 1632 samples}}} \\[1pt]
FLUX(30 steps) &1.00$\times$&5.090 & -- & -- & -- & 0.00\% \\
OC-BLO(Ours) &3.84$\times$&1.327&0.372&0.618&18.810&79.67\%\\
OC-BLO(Ours) &4.04$\times$&1.261&0.388&0.603&18.505&81.05\%\\
OC-BLO(Ours) &4.26$\times$&1.195&0.410&0.586&18.057&82.37\%\\
OC-BLO(Ours) &4.50$\times$&1.130&0.418&0.578&17.673&83.84\%\\
OC-BLO(Ours) &4.78$\times$&1.065&0.443&0.554&16.908&85.28\%\\
OC-BLO(Ours) &5.25$\times$&0.969&0.478&0.524&16.104&87.27\%\\
\hline
\end{tabular}
}
\end{minipage}
\end{table*}
\subsubsection{Quantitative Results on \textsc{FLUX.1-dev}}
\label{ex:flux-main}
We evaluate \textbf{O}nline\textbf{C}ache with \textbf{B}i\textbf{l}evel \textbf{O}ptimization (i.e., OC-BLO) on MSCOCO~\citep{DBLP:conf/eccv/LinMBHPRDZ14}, generating 800 images at a $512 \times 512$ resolution. Quantitative results are presented in Table~\ref{tab:comparison-flux} , \textit{\textbf{Block1}}. Using \textsc{FLUX.1-dev} as the baseline without caching, we compare our method against widely adopted cache-based approaches, including TeaCache and ERTACache. Under identical settings, OnlineCache demonstrates consistent performance improvements in synthesis quality and efficiency.

\subsubsection{Generalization Analysis}
\label{ex:generalization}
To assess the robustness and applicability of our proposed OnlineCache, we conduct comprehensive generalization experiments across four dimensions:
\begin{itemize}[label=\(\triangleright\), leftmargin=*, itemsep=2pt]
    \item \textbf{Cross-Resolution Generalization (\textit{Block2}):} We evaluate the policy and corrector trained at $512 \times 512$ directly on $1024 \times 1024$ generation without retraining. OnlineCache achieves a $3.25\times$ speedup, outperforming a 12-step reduction baseline ($2.46\times$) across all metrics. Compared to ERTACache, our method also attains a higher speedup while maintaining superior fidelity. This scalability is largely attributed to the policy design (Section~\ref{rl-based-decision-policy}), which utilizes global statistics (channel-wise mean and standard deviation) rather than raw spatial tensors. Since sequence length $L$ varies with image resolution, adopting such resolution-agnostic statistics avoids coupling the policy to specific sequence lengths, thereby improving generalization across different resolution.
    \item \textbf{Cross-Model Generalization (\textit{Block3}):} While direct cross-model transfer is generally constrained by architectural disparities (e.g., varying latent spaces, channel dimensions, and prediction targets), training a model-specific policy remains highly practical. The policy introduces negligible overhead (76.6MB for 24GB \textsc{FLUX.1-dev}) and trains efficiently (see Table \ref{tab:training_time_memory_cost}). Furthermore, the learned policy exhibits zero-shot transferability when underlying architectures are closely aligned. For instance, applying the \textsc{FLUX.1-dev} policy to its distilled variant, \textsc{FLUX.1-schnell} (sharing the same 3072-dimensional hidden state space), achieves a $2.31\times$ speedup with LPIPS $<0.2$.
    \item \textbf{Cross-Dataset Generalization (\textit{Block4}):} We directly evaluate the MSCOCO-trained OnlineCache on the Parti-Prompts~\citep{DBLP:journals/tmlr/YuXKLBWVKYAHHPLZBW22} dataset. The policy maintains strong performance, indicating that it captures fundamental denoising trajectory dynamics rather than dataset-specific artifacts, rendering domain-specific retraining unnecessary. Comparisons with the state-of-the-art TaylorSeer further validate the sustained efficacy of our approach under data distribution shifts.
    \item \textbf{Acceleration Ratio Generalization (\textit{Block5}):} During training with the acceleration constraint (Equation~\ref{eq:acceleration_constraint}), cache ratio converges to $\sim0.43$, yielding a $1.65\times$ speedup. At inference, the overall acceleration ratio can be flexibly controlled via arithmetic modifications to the policy's final output logits. Even under extreme acceleration (scaling from $3.84\times$ to $5.25\times$), performance degrades gracefully rather than collapsing. Specifically, LPIPS increases monotonically from $0.372$ to $0.478$, indicating robust and reliable instance-aware decisions under tight computational budgets.
\end{itemize}

\subsection{\textsc{DiT-XL/2} Experiment}
\label{ex:dit-ex}
\begin{table*}[h]
    \centering
    \caption{\textbf{Quantitative comparison on DiT-XL/2}. We compare OnlineCache against other \textbf{learning-based} acceleration approaches under diverse sampling configurations. The "FastCache baselines” are reused from the official FastCache paper~\citep{DBLP:journals/corr/abs-2505-20353} (Tables~10 and Table~12). IS and Prec denote Inception Score and Precision, respectively. \textit{Note that sFID, IS, Prec, and Recall are not reported for FastCache Baselines, as they were not evaluated in the official paper.}}
    \label{tab:comparison-dit}
    \begin{minipage}[t]{0.48\linewidth}
    \resizebox{\columnwidth}{!}{
    \begin{tabular}{lcccccc}
        \toprule
        \textbf{Method} & \textbf{FID} $\downarrow$ & \textbf{sFID} $\downarrow$ & \textbf{IS} $\uparrow$ & \textbf{Prec} $\uparrow$ & \textbf{Recall} $\uparrow$ & \textbf{Speed} $\uparrow$ \\
        \midrule
        \multicolumn{7}{c}{\textit{\textbf{FastCache Baselines}}} \\
        Baseline & 4.45 & - & - & - & - & 1.00$\times$ \\
        TeaCache & 5.09 & - & - & - & - & 1.84$\times$ \\
        AdaCache & 4.64 & - & - & - & - & 1.26$\times$ \\
        L2C      & 6.88 & - & - & - & - & 1.69$\times$ \\
        FBCache  & 4.48 & - & - & - & - & 1.63$\times$ \\
        FastCache & 4.46 & - & - & - & - & 1.74$\times$ \\
        \midrule
        \multicolumn{7}{c}{\textit{\textbf{OnlineCache with BLO (DDIM, 30 steps, CFG 1.7)}}} \\
        Baseline & 4.44 & 8.53 & 304.46 & 0.85 & 0.50 & 1.00$\times$ \\
        OC-BLO & 4.41 & 8.61 & 284.57 & 0.84 &0.50&1.54$\times$ \\
        OC-BLO & 4.43 & 8.74 & 271.91 & 0.84 &0.50&1.70$\times$ \\
        \bottomrule
    \end{tabular}
    }
    \end{minipage}\hfill
    \begin{minipage}[t]{0.49\linewidth}
    \resizebox{\columnwidth}{!}{
    \begin{tabular}{lcccccc}
        \toprule
        \textbf{Method} & \textbf{FID} $\downarrow$ & \textbf{sFID} $\downarrow$ & \textbf{IS} $\uparrow$ & \textbf{Prec} $\uparrow$ & \textbf{Recall} $\uparrow$ & \textbf{Speed} $\uparrow$ \\
        \midrule
        \multicolumn{7}{c}{\textit{\textbf{More OnlineCache Configurations}}} \\[3pt]
        \multicolumn{7}{c}{\textit{\textbf{DPM++, 50 steps, CFG 1.5}}} \\
        Baseline &3.47&9.20&268.44&0.82&0.57&1.00$\times$ \\
        OC-BLO &3.28&8.05&257.19&0.82&0.56&1.88$\times$ \\[1pt]
        \multicolumn{7}{c}{\textit{\textbf{DPM++, 25 steps, CFG 1.5}}} \\
        Baseline &3.77&10.63&264.17&0.82&0.55&1.00$\times$ \\
        OC-BLO &3.69&7.65&229.30&0.78&0.57&1.69$\times$ \\[1pt]
        \multicolumn{7}{c}{\textit{\textbf{{DPM++, 15 steps, CFG 1.5}}}} \\
        Baseline &3.95&11.09&254.78&0.82&0.55&1.00$\times$ \\
        OC-BLO &4.07&8.42&189.45&0.71&0.58&1.65$\times$ \\
        \bottomrule
    \end{tabular}
    }
    \end{minipage}
\end{table*}

Table~\ref{tab:comparison-dit} presents the results on ImageNet-256 using \textsc{DiT-XL/2}~\citep{DBLP:conf/iccv/PeeblesX23}. To comprehensively evaluate the performance of our method, we conduct comparisons against a range of widely-used methods, including training-free approaches (TeaCache~\citep{DBLP:conf/cvpr/Liu0W0QZZY025}, AdaCache~\citep{DBLP:journals/corr/abs-2411-02397}, FBCache~\citep{ParaAttention}), and \textbf{learning-based} caching methods such as L2C~\citep{DBLP:conf/nips/MaFMW24}, FastCache~\citep{DBLP:journals/corr/abs-2505-20353}. To ensure fairness, we align our settings with FastCache. Following standard protocols, we generate 50k samples across 1k classes for evaluation and ensure an \textbf{identical baseline FID}. All metrics are computed by the official OpenAI evaluator~\footnote{\url{https://github.com/openai/guided-diffusion}}. 
We further evaluate the robustness of our method across different samplers and configurations, including DDIM~\citep{DBLP:conf/iclr/SongME21} and DPM-Solver++~\citep{DBLP:journals/ijautcomp/LuZBCLZ25}. Under \textbf{\textit{DPM++, 50 steps, CFG 1.5}} setting, OnlineCache achieves a \textbf{1.88$\times$ speedup} while unexpectedly improving FID from 3.47 to 3.28. We attribute this improvement to the fact that ground-truth sampling trajectories are imperfect (see Figure~\ref{fig:visual_dit}). By selectively caching redundant steps and applying error correction, OnlineCache suppresses sampling noise and improves robustness, leading to higher-quality synthesis. Notably, these gains are achieved with \textbf{<5 hours} of additional training overhead (see Appendix~\ref{appendix2_time_memory_cost}).

\subsection{\textsc{CogVideoX-2b} Experiment}
\label{ex:cogvideox-ex}
\begin{table}[h]
\centering
\small
\caption{\textbf{Quantitative comparison on CogVideoX-2b}. We compare OnlineCache against TeaCache.}
\label{tab:comparison-cogvideox}
\resizebox{0.6\columnwidth}{!}{
\begin{tabular}{lccccc}
\hline
\addlinespace[0.1em]
\textbf{Method}
& \textbf{Speedup}~$\uparrow$ & \textbf{VBench}~$\uparrow$ & \textbf{LPIPS}~$\downarrow$ & \textbf{SSIM}~$\uparrow$ & \textbf{PSNR}~$\uparrow$ \\
\hline
\multicolumn{6}{c}{\textit{\textbf{Baselines (12 frames, 480P)}}} \\
CogVideoX & 1.00$\times$ & 78.37\% & -- & -- & -- \\
OC-BLO & 1.30$\times$ & 78.37\% & 0.019 & 0.957 & 39.849 \\
\rowcolor{green!8} TeaCache ($\lambda=0.1$)&1.31$\times$& 78.09\%& 0.062 & 0.932 & 30.309 \\
\rowcolor{green!8} OC-BLO & \textbf{1.61$\times$} & \textbf{78.23\%} & \textbf{0.057} & \textbf{0.945} & \textbf{35.705} \\
\rowcolor{lightblue} TeaCache ($\lambda=0.2$)&1.64$\times$& 77.40\%& 0.159 & 0.854 & 24.076 \\
\rowcolor{lightblue} OC-BLO & \textbf{1.79$\times$} & \textbf{78.09\%} & \textbf{0.106} & \textbf{0.921} & \textbf{29.983} \\
\bottomrule
\end{tabular}
}
\end{table}
We further extend OnlineCache to the \textbf{video generation} modality. We evaluate it on the CogVideoX-2b~\citep{DBLP:conf/iclr/YangTZ00XYHZFYZ25}.
We follow the official VBench evaluation pipeline~\citep{DBLP:conf/cvpr/HuangHYZS0Z0JCW24} and the released TeaCache implementation. For each prompt, five samples are generated using seeds 0–4. The evaluation metrics strictly adhere to the configuration used in TeaCache. To reduce computational cost, we set the number of frames to 12 and inference steps to 36, reproducing the reported TeaCache results and executing our OnlineCache under this setup. For visualization examples, see Figure~\ref{fig:visual_cogvideo}.

\subsection{Ablation Studies and Empirical Analysis}
\label{main_paper_ablation_study}
We conduct thorough ablation studies and empirical analysis to validate our system design and understand its underlying mechanisms. Specifically, we investigate:
\begin{itemize}[label=\(\triangleright\), leftmargin=*, itemsep=2pt]
    \item Policy Input States (Sec.~\ref{ablation:input_state_decision}): Ablation on the contribution of individual features in state vector.
    \item Corrector Training Targets (Sec.~\ref{appendix2_ablation_on_corrector_training_target}): Ablation on the training targets of the error corrector.
    \item Corrector Effectiveness (Sec.~\ref{appendix2_ablation_on_corrector_effectiveness}): Module-level ablation confirms the corrector's ability.
    \item Caching Dynamics (Sec.~\ref{appendix2_ablation_caching_dynamics}): An empirical analysis of the policy's adaptive caching behavior.
\end{itemize}

\subsubsection{Ablation on Policy Input States Design}
\label{ablation:input_state_decision}
We adopt channel-wise statistical summaries ($1{\times}D$) as a compact representation of latent states for policy network input and compare it with alternative designs:
\begin{itemize}[label=\(\triangleright\), leftmargin=*, itemsep=2pt]
    \item Full Latents ($L{\times}D$): Directly using raw tokens is computationally prohibitive, as sequence length $L$ can exceed 1,536 in \textsc{FLUX.1-dev}. In contrast, statistics are both efficient and informative. Moreover, this formulation is resolution-agnostic and improves generalization (see Section~\ref{ex:generalization}).
    \item Global Flattening ($1{\times}1$): Collapsing all information into a single scalar loses structural and semantic variations, making it insufficient for modeling denoising dynamics.
    \item Token-wise Statistics ($L{\times}1$): While preserving per-token information along the sequence dimension, this design discards channel-wise semantic structure. In contrast, retaining the $D$ dimension better captures feature-level semantics. This formulation is also widely adopted in practice\footnote{\url{https://github.com/AntixK/PyTorch-VAE/blob/master/models/vanilla_vae.py}}.
\end{itemize}

\begin{figure}[h]
    \centering
    \includegraphics[width=0.8\linewidth]{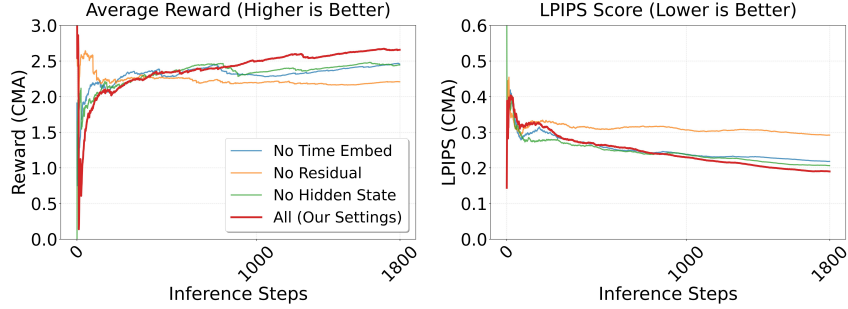}
    \caption{Ablation study on policy input state components. The full configuration (\emph{All}) achieves the highest training reward and the best LPIPS score, while removing any individual component leads to consistent performance degradation.}
    \label{fig:input_state_decision}
\end{figure}

Having adopted the $1{\times}D$ channel-wise statistical summaries, a natural question arises: \emph{which statistics should comprise it?} To assess the contribution of each component in the state vector $s_t$, we conduct an ablation study by selectively removing input features. The results in Figure~\ref{fig:input_state_decision} report Average Reward (CMA) and LPIPS across four configurations. We consider features from three sources: the cached residual $r_{\text{cached}}$, the current latent $x_t$, and the time embedding $\text{emb}_{time}(t)$. As illustrated, the full configuration (\textit{Our Settings}) consistently outperforms all ablated variants.
\begin{itemize}[label=\(\triangleright\), leftmargin=*, itemsep=2pt]
    \item Compared to \textit{No Residual:} Removing the cached statistics $r_{\text{cached}}$ results in the most severe performance degradation. Without this cached information, the policy network is unable to track error accumulation and instead relies on uninformed estimates, leading to uncontrolled collapse.
    \item Compared to \textit{No Time Embed} or \textit{No Hidden State:} 
    Removing either the time embedding $\text{emb}_{time}(t)$ or the current latent $x_t$ also causes consistent performance drops. 
    The absence of temporal information prevents stage-aware adaptation in the non-stationary diffusion denoise process, while removing the latent $x_t$ deprives the policy of the current denoising state, thereby weakening its ability to reason about subsequent denoising dynamics.
\end{itemize}
In summary, the synergy of time-awareness, current context, and cache information is essential for the policy to achieve an effective acceleration-quality trade-off.

\section{Conclusion}
We propose OnlineCache, a dynamic cache-based acceleration framework that reformulates diffusion inference as an instance-aware sequential decision-making problem. By introducing a lightweight policy network and, in its advanced variant, jointly training an error corrector under a bilevel optimization paradigm, OnlineCache adaptively navigates the delicate trade-off between generation quality and computational cost, effectively handling both sample-level and timestep-level heterogeneities. Extensive experiments demonstrate the robustness, generality, and practical effectiveness of the proposed framework. By offering a globally optimized perspective, OnlineCache establishes a principled cache-based acceleration paradigm for high-fidelity, real-time generative modeling.

\clearpage
\bibliographystyle{unsrtnat}
\bibliography{ref}


\newpage
\appendix
\vspace{0.2in}
\bigskip
\begin{center}
{\Large\bf Appendix: OnlineCache: Learning Dynamic Caching Policies with Error Correction for Efficient Diffusion Inference}
\end{center}

\noindent
\textbf{Appendix Overview.}  
This appendix is organized as follows:

\begin{itemize}[label=\(\triangleright\), leftmargin=*, itemsep=2pt]
    \item \textbf{\ref{appendix1_gradient_derivation}} presents the detailed policy gradient derivation and parameter optimization process.
    \item \textbf{\ref{appendix_related_works}} provides a more comprehensive discussion of related work.
    \item \textbf{\ref{appendix2_supplementary_experiment_details}} provides supplementary experimental details, including:
    \begin{itemize}[label=\(\triangleright\), leftmargin=*, itemsep=2pt]
        \item \textbf{~\ref{appendix2_time_memory_cost}}: Analysis of training time and memory consumption;
        \item \textbf{~\ref{appendix2_additional_ablation_study}}: Additional ablation studies beyond Section~\ref{main_paper_ablation_study} in the main paper;
        \item \textbf{~\ref{appendix2_more_experiment_results}}: Extended experimental results and detailed configuration;
        \item \textbf{~\ref{appendix2_training_dynamics}}: Training dynamics analysis on \textsc{FLUX.1-dev}, \textsc{DiT-XL/2} and \textsc{CogVideoX-2B}.
    \end{itemize}
    \item  \textbf{\ref{appendix3_limitations}} discusses the limitations of the paper.
    \item  \textbf{\ref{appendix4_broader_impacts}} provides a discussion of broader impacts.
    \item  \textbf{\ref{appendix5_visual_results}} includes a three-page full visual demonstration showcasing representative generation results.
\end{itemize}

\vspace{1\baselineskip}
\section{Policy Gradient derivation and Parameter Optimization Process}
\label{appendix1_gradient_derivation}

In this section, we provide the mathematical derivation for the optimization of the policy network parameters $\theta$. Since the caching decision $a_t$ is discrete and sampled from a Bernoulli distribution, the operation is non-differentiable with respect to $\theta$. Therefore, standard backpropagation cannot be applied directly through the action $a_t$ to the reward $R(\tau)$. Instead, we utilize the REINFORCE algorithm (Monte Carlo Policy Gradient) to estimate the gradient of the expected reward.

\subsection{Policy Gradient of the Objective Function}
Our primary goal is to maximize the expected cumulative reward $J(\theta)$ over trajectories $\tau$:
\begin{equation}
    J(\theta) = \mathbb{E}_{\tau \sim \pi_\theta} [R(\tau)] = \sum_{\tau} P(\tau \mid \theta) R(\tau),
\end{equation}
where $P(\tau \mid \theta)$ is the probability of a trajectory $\tau = \{(s_0, a_0), \dots, (s_T, a_T)\}$ occurring under policy $\pi_\theta$. The state are determined by the diffusion dynamics and only actions depend on $\theta$, the probability of a trajectory is given by:
\begin{equation}
    P(\tau \mid \theta) \propto \prod_{t=1}^T \pi_\theta(a_t \mid s_t).
\end{equation}

To maximize $J(\theta)$, we perform gradient ascent. $\nabla_\theta J(\theta)$ is derived using the log-derivative trick:
\begin{equation}
\begin{aligned}
    \nabla_\theta J(\theta) &= \nabla_\theta \sum_{\tau} P(\tau \mid \theta) R(\tau) = \sum_{\tau} R(\tau) \nabla_\theta P(\tau \mid \theta)\\
    &= \sum_{\tau} R(\tau) P(\tau \mid \theta) \nabla_\theta \log P(\tau \mid \theta) = \sum_{\tau} R(\tau) P(\tau \mid \theta) \nabla_\theta \log \prod_{t=1}^T \pi_\theta(a_t \mid s_t) \\
    &= \sum_{\tau} P(\tau \mid \theta) R(\tau) \sum_{t=1}^T \nabla_\theta \log \pi_\theta(a_t \mid s_t) = \mathbb{E}_{\tau \sim \pi_\theta} \left[ R(\tau) \sum_{t=1}^T \nabla_\theta \log \pi_\theta(a_t \mid s_t) \right].
\end{aligned}
\end{equation}
In practice, we approximate this expectation using a batch of $B$ sampled trajectories.

\subsection{Empirical Analysis: Variance Reduction via Batch Normalization}
\label{sec:batch_norm_analysis}
\begin{figure}[h]
    \centering
    \includegraphics[width=1.0\linewidth]{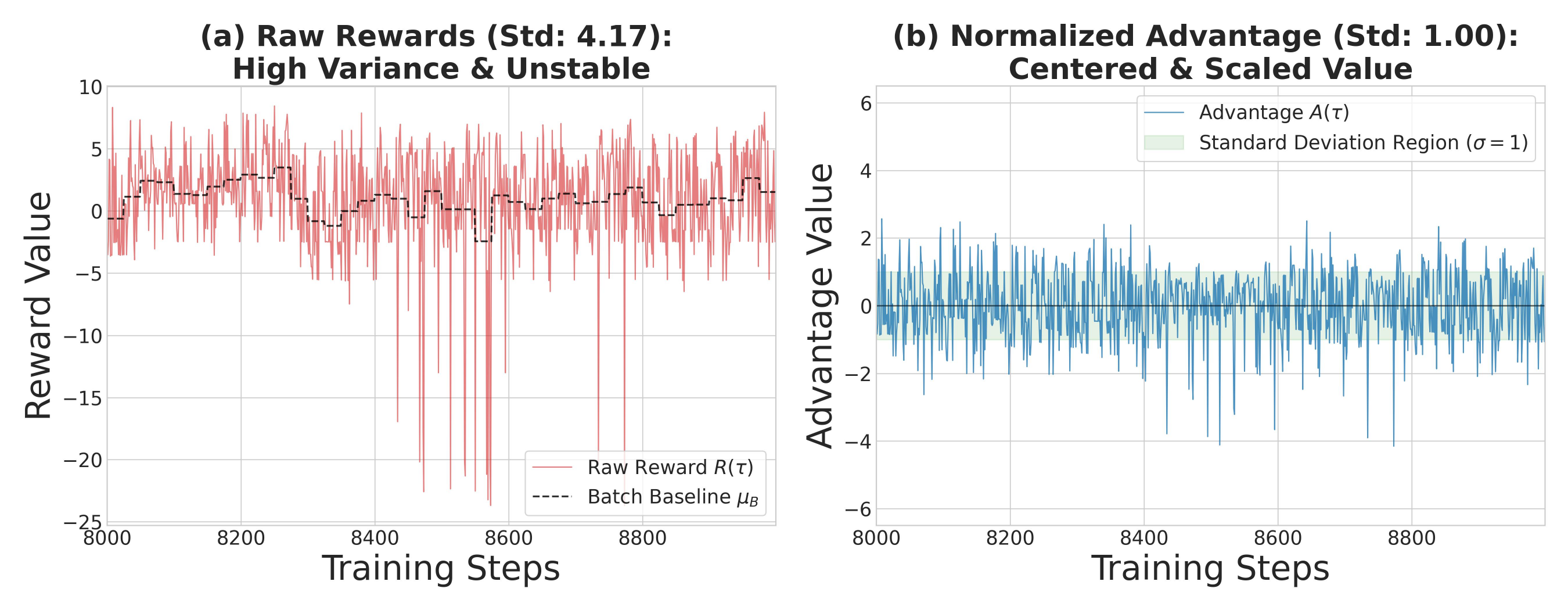}
    \caption{Comparison of Raw Rewards and Normalized Advantages. (a) The raw reward $R(\tau)$ (red) exhibits high variance, even when the batch baseline $\mu_B$ (black dashed) is relatively stable. (b) The normalized advantage $A(\tau)$ (blue) effectively centers the signal (mean 0, std 1), ensuring that gradient updates are driven by \textbf{relative performance within the batch}.}
    \label{fig:reward_variance}
\end{figure}

To demonstrate the necessity of reward normalization, we visualize the training dynamics of the policy network (based on the \textsc{CogVideoX-2b}~\citep{DBLP:conf/iclr/YangTZ00XYHZFYZ25} model) during the optimizing iterations from 8k to 9k. At this stage, the policy network approaches convergence, exhibiting relatively strong stability. However, the raw reward signals still remain highly stochastic. We present the comparison between raw rewards and normalized advantages in Figure~\ref{fig:reward_variance}.

\textbf{Instability of Raw Rewards.} As shown in Figure~\ref{fig:reward_variance} (a), the raw reward signal $R(\tau)$ (red line) is characterized by high-frequency oscillations and significant variance (std: 4.17). Although the batch baseline $\mu_B$ (black dashed line, computed with batch size $B=25$) captures the local trend, the individual trajectories fluctuate wildly around this baseline. Since the model is near convergence, $\mu_B$ shows no obvious upward trend. Directly optimizing against such raw signals can introduce substantial noise into the policy gradient, potentially destabilizing the learning process.

\textbf{Stability via Normalized Advantage.} To address this, we standardize rewards within each batch to compute the advantage $A(\tau_i)$, normalization is defined in Equation~\ref{eq:advantage_norm}. Figure~\ref{fig:reward_variance} (b) illustrates the resulting advantage reward signal. This transformation provides two key benefits:
\begin{enumerate}[label=\textbf{(\arabic*)}, leftmargin=*, itemsep=2pt]
    \item \textbf{Relative Performance Quantification:} By subtracting the batch mean $\mu_R$ and dividing by the standard deviation $\sigma_R$, the advantage $A(\tau_i)$ captures the \textit{relative merit} of a trajectory compared to its peers within the same batch. A positive value (i.e., $A(\tau_i) > 0$) indicates above-average performance, providing a stable, scale-invariant optimization signal.
    \item \textbf{Gradient Stabilization:} The normalization constrains the advantage values (with unit variance, typically within $[-2, 2]$), as visualized by the shaded region in Figure~\ref{fig:reward_variance} (b). This mitigates the influence of reward outliers on gradient magnitude, leading to more stable convergence.
\end{enumerate}

\subsection{Entropy Regularization}
To prevent the policy from collapsing into a deterministic solution too early (e.g., always caching or never caching), we augment the objective with an \textbf{entropy regularization} term. As defined in Equation \ref{eq:entropy_term}. Maximizing this entropy encourages the predicted probabilities $p_t$ to stay closer to $0.5$ when the reward signal is ambiguous, thereby promoting exploration of different caching strategies. 
Importantly, unlike the discrete action sampling, $\mathcal{H}$ is a differentiable function of the network output $p_t$, allowing gradients to flow directly through the policy network without Monte Carlo estimation.

\subsection{Surrogate Loss and Parameter Update}
\label{sec:surrogate_loss}
We design a tailored \textbf{surrogate loss} to update the policy gradient within standard deep learning frameworks. By combining the \textbf{policy objective} and \textbf{entropy regularization}, we define the total surrogate loss $\mathcal{L}(\theta)$ for a batch of $B$ trajectories as:
\begin{equation}
\label{eq:appendix_total_loss}
    \mathcal{L}(\theta) = \frac{1}{B} \sum_{i=1}^B \left[ \underbrace{- A(\tau_i) \sum_{t=1}^T \log \pi_\theta(a_{i,t} \mid s_{i,t})}_{\mathcal{L}_{\text{policy}}} \underbrace{- \lambda_e \frac{1}{T} \sum_{t=1}^T \mathcal{H}(\pi_\theta(s_{i,t}))}_{\mathcal{L}_{\text{entropy}}} \right].
\end{equation}
Note that the negative signs are introduced to convert the maximization problem (maximizing Reward and Entropy) into a minimization problem suitable for standard gradient descent optimizers.

\subsubsection{Gradient Flow Mechanism}
The optimization proceeds by computing the gradient $\nabla_\theta \mathcal{L}(\theta)$. The two components of the loss contribute to the parameter update in distinct ways:

\begin{enumerate}[label=\textbf{(\arabic*)}, leftmargin=*, itemsep=2pt]
    \item \textbf{Policy Gradient Flow ($\mathcal{L}_{\text{policy}}$):} The term $A(\tau_i)$ is treated as a detached scalar (fixed constant) during backpropagation. The gradient flows only through the log-probability term:
    \begin{equation}
        \nabla_\theta \mathcal{L}_{\text{policy}} = - \frac{1}{B} \sum_{i=1}^B \sum_{t=1}^T A(\tau_i) \cdot \nabla_\theta \log \pi_\theta(a_{i,t} \mid s_{i,t}).
    \end{equation}
    This term implements the intuition of REINFORCE:
    \begin{itemize}[label=\(\triangleright\), leftmargin=*, itemsep=2pt]
        \item We do not differentiate through the discrete action sampling $a_t$ (from Bernoulli distribution, which is non-differentiable). Instead, we differentiate the \textit{probability} of the sampling action.
        \item If the advantage $A(\tau_i)$ is positive, the negative sign ensures the gradient descent step \textit{increases} the log-probability of that action; if negative, conversely, when negative, it \textit{decreases} the log-probability, thereby discouraging that action.
    \end{itemize}

    \item \textbf{Entropy Gradient Flow ($\mathcal{L}_{\text{entropy}}$):} The entropy $\mathcal{H}$ is a deterministic and differentiable function of the network output distribution, allowing gradients to propagate directly via the chain rule:
    \begin{equation}
        \nabla_\theta \mathcal{L}_{\text{entropy}} = - \frac{\lambda_e}{B \cdot T} \sum_{i=1}^B \sum_{t=1}^T \frac{\partial \mathcal{H}}{\partial \pi_\theta} \cdot \nabla_\theta \pi_\theta(s_{i,t}).
    \end{equation}
    This standard backpropagation encourages the policy to output probability distributions with higher variance, preventing mode collapse (model chooses to cache all or compute all).
\end{enumerate}

\subsubsection{Parameter Update}
Finally, the parameters are updated iteratively using the Adam optimizer with learning rate $\eta$:
\begin{equation}
    \nabla_\theta \mathcal{L}(\theta) = \nabla_\theta \mathcal{L}_{\text{policy}} + \nabla_\theta \mathcal{L}_{\text{entropy}}, 
    \quad\theta \leftarrow \theta - \eta \left( \nabla_\theta \mathcal{L}(\theta) \right).
\end{equation}

\vspace{1\baselineskip}
\section{Comprehensive Related Works}
\label{appendix_related_works}
\paragraph{Model- and Solver-based Acceleration.}
Substantial works have been devoted to diffusion inference acceleration. Existing approaches primarily focus on optimizing the model architecture or the representation. Key strategies include \textbf{pruning}~\citep{DBLP:conf/nips/FangMW23, DBLP:conf/cvpr/CastellsSKC22, DBLP:journals/corr/abs-2404-11098}, \textbf{quantization}~\citep{DBLP:conf/cvpr/ShangYXW023, DBLP:conf/nips/SoLAKP23, DBLP:conf/nips/HeLLWZZ23, DBLP:conf/nips/LiX0S023}, and \textbf{distillation}~\citep{DBLP:journals/corr/abs-2101-02388, DBLP:conf/iclr/SalimansH22, DBLP:conf/cvpr/YangZFW23, DBLP:journals/corr/abs-2311-17042}. In parallel, significant advances have been made in \textbf{sampling solvers}~\citep{DBLP:journals/corr/abs-2105-14080, DBLP:conf/iclr/SongME21, DBLP:conf/nips/0011ZB0L022, DBLP:conf/iclr/BaoLZZ22, DBLP:conf/iclr/Liu0LZ22, DBLP:conf/nips/KarrasAAL22, 
DBLP:conf/iclr/ZhangTC23, DBLP:conf/iclr/ZhangC23, DBLP:journals/ijautcomp/LuZBCLZ25} to accelerate denoising. 

\paragraph{Cache-based Acceleration.}
Cache-based acceleration exploits temporal redundancy between the current step and historical denoising steps to speedup inference without altering the backbone model. 
Previous works mainly adopt \textbf{static caching} schedules or utilize architectural properties. DeepCache~\citep{DBLP:conf/cvpr/MaFW24} and FORA~\citep{DBLP:journals/corr/abs-2407-01425} perform periodic feature reuse with fixed intervals for U-Net and DiT, respectively. To mitigate drift from direct reuse, subsequent methods introduce statistical calibration (Block Caching~\citep{DBLP:conf/cvpr/WimbauerWSDHHSZ24}), increments caching ($\Delta$-DiT~\citep{DBLP:journals/corr/abs-2406-01125}) or error rectification (ERTACache~\citep{DBLP:journals/corr/abs-2508-21091}). Others exploit architectural redundancies like reusing U-Net encoder features~\citep{DBLP:conf/nips/LiH0KLLYWC024}, broadcasting attention output~\citep{DBLP:conf/iclr/ZhaoJW025} and caching for window attention~\citep{DBLP:conf/nips/YuanZPNZZYD024}.
Recognizing that static methods fail to capture instance specificity, many studies shift towards \textbf{dynamic strategies}. Methods like TeaCache~\citep{DBLP:conf/cvpr/Liu0W0QZZY025} and DiCache~\citep{DBLP:journals/corr/abs-2508-17356} utilize lightweight indicators (timestep embedding differences and shallow feature probes, respectively) to dynamically trigger computation. Similarly, in video modality, AdaCache~\citep{DBLP:journals/corr/abs-2411-02397} and EasyCache~\citep{DBLP:journals/corr/abs-2507-02860} adjust caching policies based on motion complexity and transformation vectors. 
Moving beyond simple reuse, forecasting methods aim to predict future features. TaylorSeer~\citep{DBLP:journals/corr/abs-2503-06923} uses Taylor expansion to extrapolate feature evolution, SpeCa~\citep{DBLP:journals/corr/abs-2509-11628} adopts a forecast-then-verify paradigm inspired by speculative decoding. Furthermore, optimization has extended to finer granularities. Token-level methods~\citep{DBLP:conf/iclr/ZouLLHZ25, DBLP:journals/corr/abs-2509-10312} reduce redundancy by selectively processing informative tokens or compressing attention. 
Finally, distinct from heuristic rules, \textbf{learning-based methods} often train routers~\citep{DBLP:conf/nips/MaFMW24} or linear compensation layers~\citep{DBLP:journals/corr/abs-2505-20353} to optimize caching decisions and approximate cached features. These approaches are not training-free but consistently achieve the state-of-the-art performance reported to date. Our proposed method, \textbf{OnlineCache}, follows a similar learning-based paradigm.

\paragraph{Bilevel Optimization.}
Bilevel optimization (BLO), which involves nesting one optimization problem within the constraints of another, serves as a fundamental mathematical framework for hierarchical optimizing tasks.~\citep{DBLP:journals/ior/BrackenM73a} first formalized the general class of mathematical programs with optimization problems in the constraints, establishing the theoretical bedrock for this field. In the context of modern machine learning and deep learning, bilevel optimization has been widely adopted to model learning problems with hierarchical structures~\citep{DBLP:journals/corr/abs-2308-00788}, such as hyperparameter optimization~\citep{DBLP:conf/icml/FranceschiFSGP18, DBLP:conf/aistats/ShabanCHB19}, model pruning~\citep{DBLP:conf/nips/ZhangYR0CHW022} and model-agnostic meta-learning~\citep{DBLP:conf/icml/FinnAL17,
DBLP:conf/nips/RajeswaranFKL19}.

\vspace{1\baselineskip}
\section{Supplementary Experiment Details}
\label{appendix2_supplementary_experiment_details}
\subsection{Analysis of Training Time and Memory Consumption}
\label{appendix2_time_memory_cost}
\begin{table}[h]
    \centering
    \caption{Training cost across different models. We report the number of training samples, total training time (including both policy and corrector stages), and peak GPU memory usage. All measurements are conducted on a single A100-80GB GPU.}
    \label{tab:training_time_memory_cost}
    \resizebox{\columnwidth}{!}{
    \begin{tabular}{lccccc}
        \toprule
        \textbf{Model} & \textbf{Training Samples} & \textbf{Bilevel Iterations} & \textbf{Time per Iteration} & \textbf{Total Time} & \textbf{Peak GPU Memory} \\
        \midrule
        \textsc{FLUX.1-dev} & 1k  & 4 / 1 & $\sim$43.1min / $\sim$49.7min & $\sim$25.9h & $\sim$38.0GB \\
        \textsc{DiT-XL/2} & 2.5k & 4 / 1 & $\sim$27.8min / $\sim$30.0min & $\sim$4.7h & $\sim$4.0GB \\
        \textsc{CogVideoX-2b} & 200 & 4 / 1 & $\sim$36.6min / $\sim$36.1min & $\sim$48.7h & $\sim$29.0GB \\
        \bottomrule
    \end{tabular}
    }
\end{table}

\subsubsection{Analysis of Computational Time Cost}
Table \ref{tab:training_time_memory_cost} details the computational overhead of OnlineCache. While it requires an additional training phase—unlike traditional training-free caching methods—this one-time cost is well justified by the resulting empirical gains. As shown in our experiments, OnlineCache consistently achieves a state-of-the-art speed-quality trade-off across diverse configurations. Notably, this gain incurs only a modest one-time training cost: the policy for \textsc{FLUX.1-dev} converges within 26 hours on a single A100 GPU, making training practically affordable. Moreover, the learned policy exhibit strong zero-shot generalization ability (see Section~\ref{ex:generalization}), further supporting the effectiveness.

\subsubsection{Memory-Efficient Implementation of Bilevel Training}
Implementing the bilevel optimization framework described in Section \ref{bilevel-optimization-framework} presents significant computational challenges. A standard naive backpropagation through the time (BPTT) for a diffusion process with $T=30$ steps requires storing the computation graph for the entire unrolling. For high-resolution latent diffusion models such as \textsc{FLUX.1-dev}, this results in memory consumption exceeding 400GB, which is prohibitive for standard hardware configurations. To address this, we implement a memory-efficient training strategy equivalent to \textbf{Truncated Backpropagation Through Time (TBPTT)} with a horizon of $k=1$ (Step-wise Optimization). With these techniques, GPU memory usage during training can be effectively controlled within an acceptable range (see Table~\ref{tab:training_time_memory_cost}).

In our implementation, this memory-efficient training scheme is realized via a step-wise optimization procedure that decouples temporal gradient dependencies across steps. The details are as follows:
\begin{enumerate}[label=\textbf{(\arabic*)}, leftmargin=*, itemsep=2pt]
    \item \textbf{Input Detachment:} At the beginning of step $t$, the input latent $x_t$ is detached from the computation graph: $x_t \leftarrow x_t.\texttt{detach()}$. This blocks gradient flow to previous timesteps.
    \item \textbf{Selective Gradient Activation:} We adopt a manually fragmented design to reduce memory usage, enabling gradients (\texttt{torch.set\_grad\_enabled(True)}) only for stage-specific modules, so that inner- and outer-loop optimizations activate gradients only for their respective components.
    \item \textbf{Immediate Backward Pass:} The outer and inner losses are computed and backpropagated immediately after the forward pass at step $t$. Gradients are accumulated in the optimizer, and the computation graph for step $t$ is promptly released before proceeding to $t-1$.
\end{enumerate}

\subsection{Additional Ablation Studies and Empirical Analysis}
\label{appendix2_additional_ablation_study}

\subsubsection{Ablation on Corrector Training Target}
\label{appendix2_ablation_on_corrector_training_target}

\begin{figure}[h]
    \centering
    \includegraphics[width=\columnwidth]{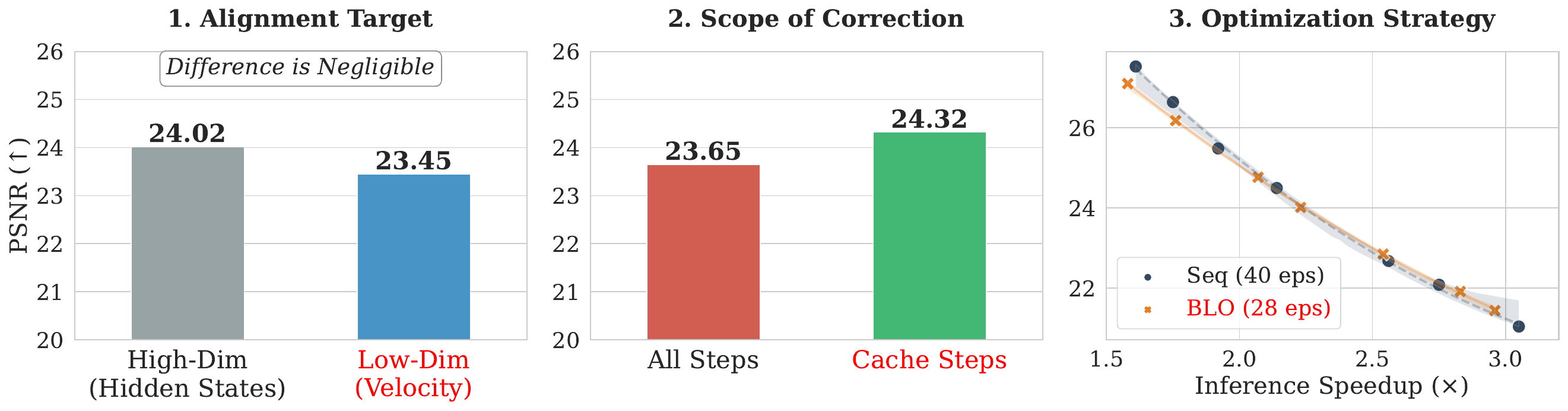}
    \caption{Ablation on corrector training. \textbf{Left}: Alignment target (High-dim vs. Low-dim). \textbf{Middle}: Scope of correction (All steps vs. Cache steps). \textbf{Right}: Optimization strategy (Sequential vs. BLO). Marked in red is the strategy we chose.}
    \label{fig:ablation_study}
\end{figure}
The right panel of Table \ref{tab:appendix2_experiment_results} investigates three critical design choices in our framework, with the ablation results visualization shown in Figure \ref{fig:ablation_study} for easier understanding. 
\begin{enumerate}[label=\textbf{(\arabic*)}, leftmargin=*, itemsep=2pt]
\item \textbf{Alignment Target: High-dim vs. Low-dim.}
We compare aligning the high-dimension \textit{hidden\_states} against the final low-dimensional output (normalized and projected transformer output velocity). Empirical results show negligible performance difference between the two strategies. However, aligning the low-dimensional velocity allows for a significantly more lightweight MLP-based corrector design, reducing inference overhead. Consequently, we adopt velocity alignment for its efficiency. As shown in the Figure \ref{fig:ablation_study} (left), we choose to align the output \textit{velocity} (64 dimensions in \textsc{FLUX.1-dev}~\citep{DBLP:journals/corr/abs-2506-15742} model, compared to 3072 dimensions of the \textit{hidden\_states}) and design the corrector to be very lightweight. During inference, adding the policy and corrector incurs a \textbf{negligible additional inference time cost}.
\item \textbf{Scope of Correction: All Steps vs. Cache Steps.} 
We investigated whether the corrector should be applied at every denoising step or only at the cached steps. Our results indicate that restricting correction \textit{to cached steps only} yields better performance. This can be attributed to the fact that the initial diffusion steps—typically not cached by the policy network—are critical for establishing semantic content. Since the baseline model already produces accurate outputs at these early steps, introducing a corrector could disrupt the trajectory and accumulate errors. Limiting the correction to cached steps also reduces the computational cost during training.
\item \textbf{Optimization Strategy: Sequential vs. Bilevel Optimization.} 
We compare our BLO strategy with sequential training, where the policy is trained to convergence first, followed by the corrector.
\begin{itemize}[label=\(\triangleright\), leftmargin=*, itemsep=2pt]
    \item \textbf{Sequential Training:} Typically requires $\sim$40 epochs to converge the policy, followed by >10 additional epochs to train the error corrector, resulting in a decoupled optimization process that fails to achieve genuine joint coordination between the two modules. 
    \item \textbf{Bilevel Optimization:} The corrector continuously adapts to the evolving policy, enabling stable training and faster convergence. Empirically, our method reaches joint convergence in just 28 outer epochs and 7 inner epochs while achieving comparable inference quality. 
\end{itemize}
\end{enumerate}

\subsubsection{Ablation on Corrector Effectiveness}
\label{appendix2_ablation_on_corrector_effectiveness}
\begin{figure}[h]
    \centering
    \includegraphics[width=\linewidth]{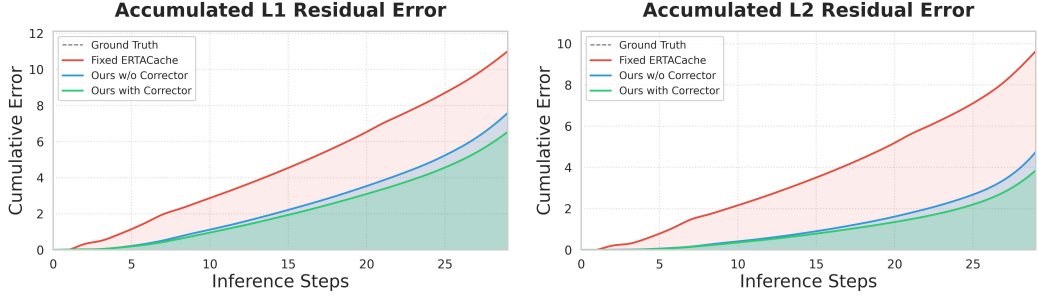}
    \caption{Accumulated $L_1$ and $L_2$ errors under different methods. The plot shows cumulative error over 30 inference steps, averaged across 50 generated samples using the \textsc{FLUX.1-dev} model.}
    \label{fig:ablation_error_accumulation}
\end{figure}
Figure~\ref{fig:ablation_error_accumulation} illustrates the accumulated $L_1$ and $L_2$ residual error for ERTACache~\citep{DBLP:journals/corr/abs-2508-21091} and our proposed OnlineCache (vanilla and bilevel optimization version) over a 30-step inference trajectory. 
Without the Corrector, residual errors in OnlineCache exhibit progressive accumulation over timesteps, reflecting the compounding effect of cache-induced approximation errors during long-range sampling. Although this accumulation is initially mild, the drift becomes increasingly pronounced as inference proceeds. 
By contrast, the inclusion of the Corrector consistently suppresses this error growth, maintaining a uniformly lower error profile throughout the entire trajectory. This demonstrates that the Corrector effectively compensates for cache-induced residuals and prevents small local deviations from amplifying into significant global errors over time.

\subsubsection{Analysis of Caching Dynamics}
\label{appendix2_ablation_caching_dynamics}
\begin{figure}[h]
    \centering
    \includegraphics[width=0.44\linewidth]{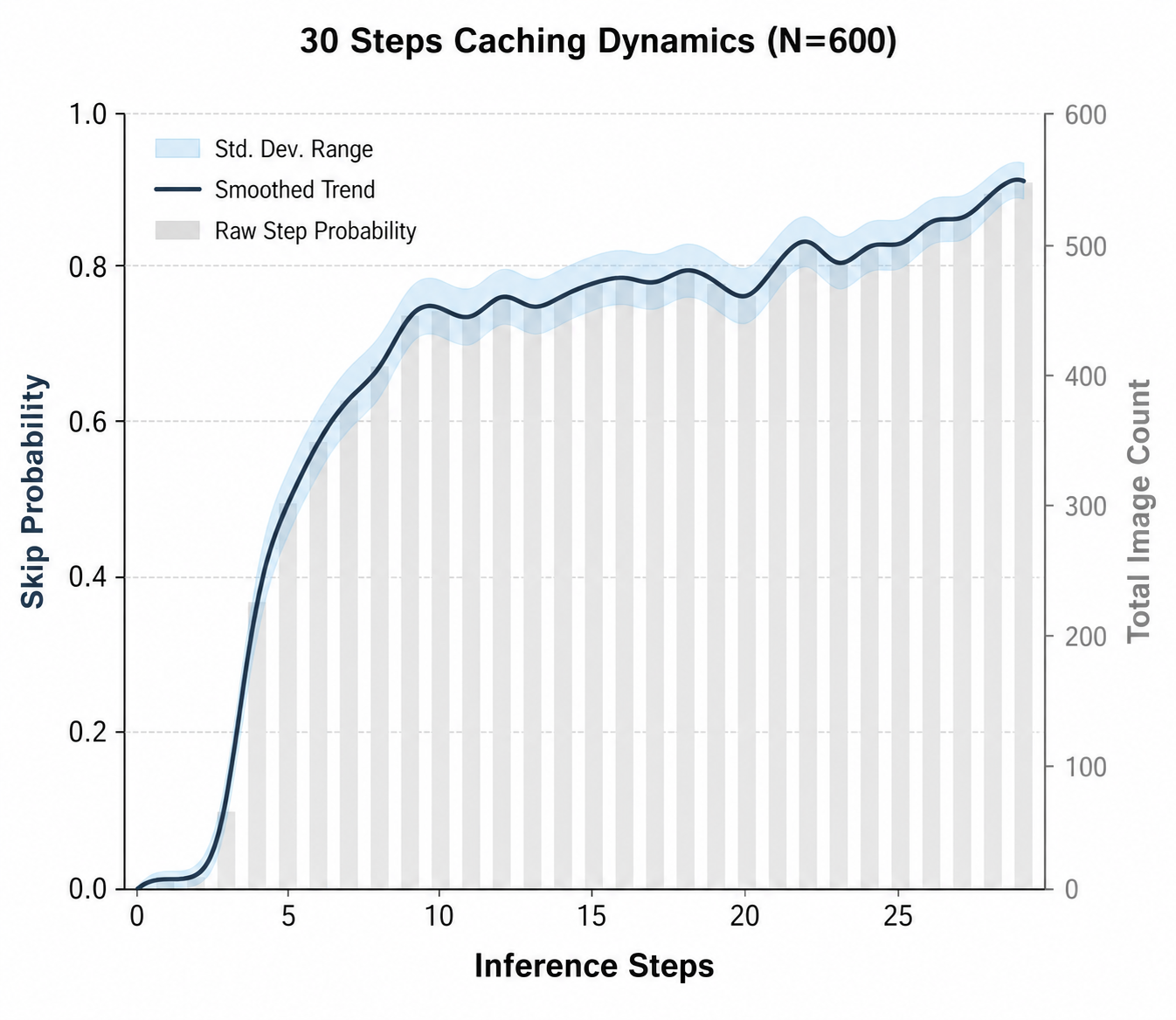}\hfill
    \includegraphics[width=0.55\linewidth]{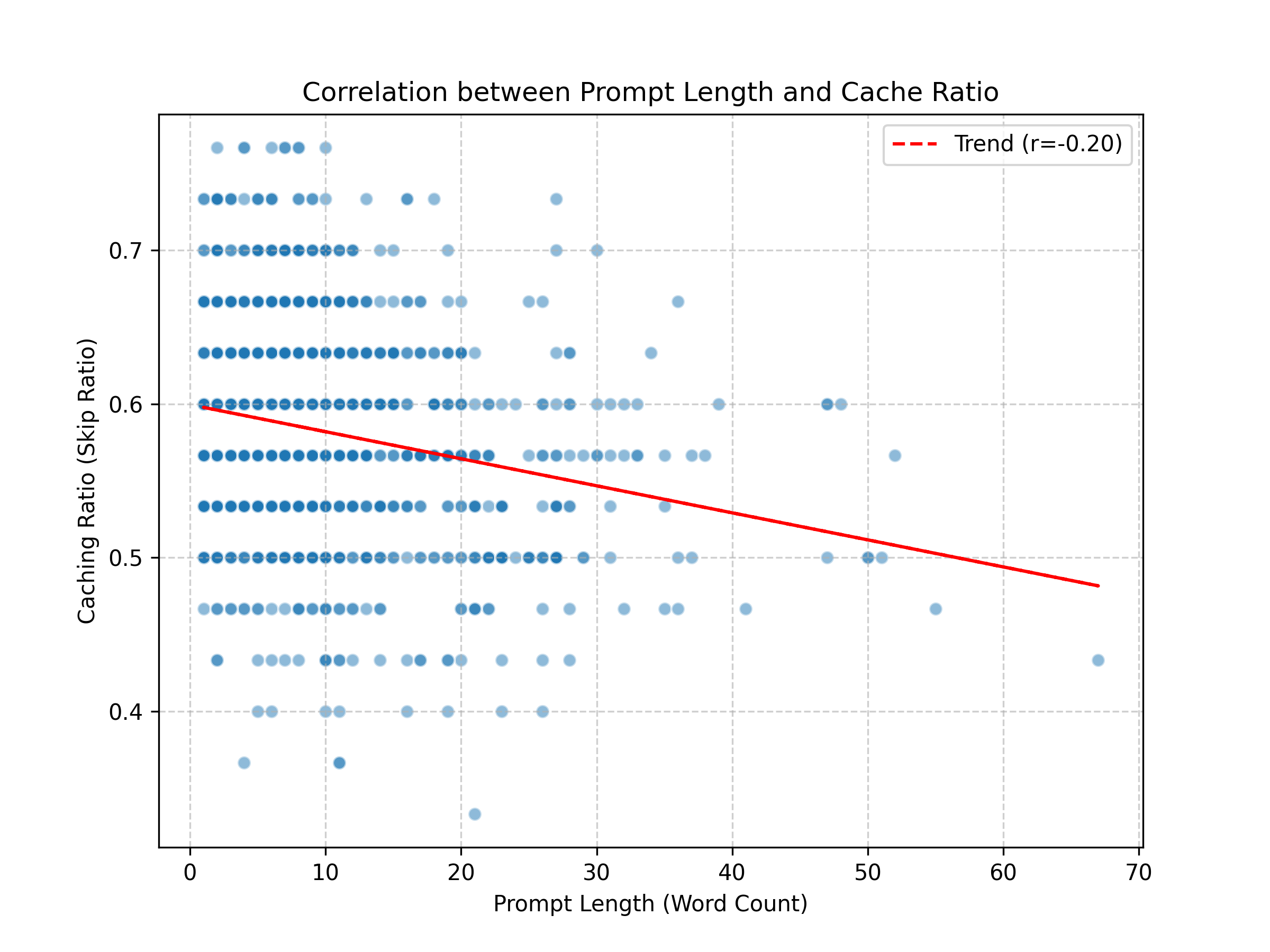}
    \caption{Adaptive caching dynamics of OnlineCache. \textbf{Left:} Average step-wise caching frequency evaluated on MSCOCO. \textbf{Right:} Correlation between prompt complexity (proxied by word count) and the cache ratio, evaluated on Parti-Prompts.}
    \label{fig:ablation_caching_dynamics}
\end{figure}
After training, the learned policy network exhibits strong adaptivity at both the timestep and sample levels, as evidenced by the following observations:
\begin{enumerate}[label=\textbf{(\arabic*)}, leftmargin=*, itemsep=2pt]
    \item \textbf{Timestep-Level Adaptivity.} In Figure~\ref{fig:ablation_caching_dynamics} (Left), the step-wise caching frequency exhibits a distinct pattern: conservative caching in early timesteps and more aggressive in later ones. This behavior inherently aligns with a fundamental property of diffusion generation: early denoising steps involve large inter-step transitions that establish global structure, making them highly sensitive to approximation errors. In contrast, later steps primarily perform fine-grained refinements with smaller updates, where caching introduces minimal perceptual degradation~\citep{DBLP:journals/corr/abs-2508-21091}.
    \item \textbf{Sample-Level Adaptivity.} To investigate prompt-aware adaptivity, we evaluate the learned policy on the Parti-Prompts dataset~\citep{DBLP:journals/tmlr/YuXKLBWVKYAHHPLZBW22}. Using \textit{word count} as a proxy for prompt complexity, Figure~\ref{fig:ablation_caching_dynamics} (Right) reveals a statistically significant negative correlation between prompt complexity and the assigned cache ratio, with a Pearson correlation coefficient of $r=-0.1994$. The fitted regression line demonstrates a clear downward trend, confirming that the policy dynamically reduces caching (allocating more computational steps) when processing more complex prompts, thereby preserving generation fidelity for difficult instances.
\end{enumerate}

\subsection{Extended experimental results and detailed configuration}
\label{appendix2_more_experiment_results}
\begin{table*}[h]
\small
\caption{\textbf{Extended experimental results on \textsc{FLUX.1-dev}.} The \textbf{left panel} shows the training dynamics of the vanilla OnlineCache across different checkpoints, revealing underfitting, optimal, and overfitting regimes. The \textbf{right panel} presents ablations on the corrector's alignment target, the scope of correction, and sequential optimization versus our BLO strategy.}
\label{tab:appendix2_experiment_results}
\begin{minipage}[t]{0.50\columnwidth}
\centering
\resizebox{\columnwidth}{!}{
\begin{tabular}{lcccccc}
\hline
\multirow{2}{*}{\textbf{Method}} 
& \multicolumn{2}{c}{\textbf{Efficiency}} 
& \multicolumn{3}{c}{\textbf{Visual Quality}} 
& \multirow{2}{*}{\textbf{CR}~$\uparrow$} \\
\cline{2-6}
& \textbf{Speed}~$\uparrow$ 
& \textbf{Lat}~$\downarrow$ 
& \textbf{LPIPS}~$\downarrow$ 
& \textbf{SSIM}~$\uparrow$ 
& \textbf{PSNR}~$\uparrow$ 
& \\ \hline
\multicolumn{7}{c}{\textit{\textbf{Previous Baselines}}} \\[2pt]
FLUX.1-dev  & 1.00x & 5.099 & --   & --   & --   & 0.00\% \\
\addlinespace[0.3em]
\hline
\multicolumn{7}{c}{\textit{\textbf{Vanilla OnlineCache}}} \\[2pt]
\textbf{Underfitting}&&&&&&\\
OnlineCache(12k) & 2.56x & 1.991 & 0.253 & 0.719 & 19.934 & 65.26\% \\
OnlineCache(12k) & 2.21x & 2.306 & 0.207 & 0.763 & 21.026 & 58.40\% \\
OnlineCache(12k) & 1.84x & 2.760 & 0.162 & 0.809 & 22.606 & 48.80\% \\
OnlineCache(20k) & 2.88x & 1.772 & 0.252 & 0.719 & 20.477 & 69.62\% \\
OnlineCache(20k) & 2.49x & 2.045 & 0.205 & 0.763 & 21.529 & 63.19\% \\
OnlineCache(36k) & 3.20x & 1.596 & 0.300 & 0.675 & 19.755 & 74.09\% \\
OnlineCache(36k) & 3.02x & 1.691 & 0.276 & 0.697 & 20.173 & 72.07\% \\
\addlinespace[0.3em]
\textbf{Best Range}&&&&&&\\
OnlineCache(38k) & 2.97x & 1.718 & 0.255 & 0.716 & 20.903 & 71.21\% \\
OnlineCache(38k) & 2.74x & 1.860 & 0.222 & 0.747 & 21.591 & 67.84\% \\
OnlineCache(38k) & 2.51x & 2.033 & 0.195 & 0.771 & 22.200 & 65.28\% \\
OnlineCache(38k) & 2.13x & 2.394 & 0.137 & 0.828 & 23.955 & 56.61\% \\
OnlineCache(38k) & 1.84x & 2.770 & 0.106 & 0.861 & 25.251 & 49.37\% \\
\addlinespace[0.3em]
OnlineCache(40k) & 3.00x & 1.702 & 0.256 & 0.711 & 20.935 & 71.83\% \\
OnlineCache(40k) & 2.88x & 1.770 & 0.247 & 0.720 & 21.159 & 70.85\% \\
OnlineCache(40k) & 2.74x & 1.863 & 0.220 & 0.746 & 21.609 & 68.55\% \\
OnlineCache(40k) & 2.62x & 1.945 & 0.202 & 0.762 & 22.145 & 66.95\% \\
OnlineCache(40k) & 2.56x & 1.993 & 0.195 & 0.771 & 22.243 & 65.90\% \\
OnlineCache(40k) & 2.47x & 2.064 & 0.182 & 0.783 & 22.647 & 64.69\% \\
OnlineCache(40k) & 2.36x & 2.161 & 0.167 & 0.797 & 23.086 & 62.28\% \\
OnlineCache(40k) & 2.24x & 2.274 & 0.159 & 0.806 & 23.232 & 60.36\% \\
OnlineCache(40k) & 2.14x & 2.381 & 0.140 & 0.826 & 23.981 & 58.22\% \\
OnlineCache(40k) & 2.05x & 2.488 & 0.129 & 0.837 & 24.420 & 55.45\% \\
\addlinespace[0.3em]
\textbf{Overfitted}&&&&&&\\
OnlineCache(42k) & 3.04x & 1.677 & 0.272 & 0.699 & 20.296 & 72.50\% \\
OnlineCache(42k) & 2.89x & 1.765 & 0.253 & 0.716 & 20.680 & 70.88\% \\
OnlineCache(42k) & 2.83x & 1.804 & 0.237 & 0.728 & 21.000 & 69.62\% \\
OnlineCache(42k) & 2.64x & 1.931 & 0.220 & 0.746 & 21.524 & 67.63\% \\
OnlineCache(45k) & 3.08x & 1.656 & 0.273 & 0.698 & 20.276 & 72.93\% \\
OnlineCache(45k) & 2.98x & 1.710 & 0.262 & 0.710 & 20.550 & 71.76\% \\
OnlineCache(50k) & 2.97x & 1.717 & 0.270 & 0.702 & 20.362 & 71.68\% \\
OnlineCache(50k) & 2.72x & 1.877 & 0.230 & 0.740 & 21.144 & 68.27\% \\
\hline
\end{tabular}
}
\end{minipage}
\hfill
\begin{minipage}[t]{0.49\columnwidth}
\centering
\resizebox{\columnwidth}{!}{
\begin{tabular}{lcccccc}
\hline
\multirow{2}{*}{\textbf{Method}} 
& \multicolumn{2}{c}{\textbf{Efficiency}} 
& \multicolumn{3}{c}{\textbf{Visual Quality}} 
& \multirow{2}{*}{\textbf{CR}~$\uparrow$} \\
\cline{2-6}
& \textbf{Speed}~$\uparrow$ 
& \textbf{Lat}~$\downarrow$ 
& \textbf{LPIPS}~$\downarrow$ 
& \textbf{SSIM}~$\uparrow$ 
& \textbf{PSNR}~$\uparrow$ 
& \\ \hline
\multicolumn{7}{c}{\textit{\textbf{OnlineCache with Corrector Post-Training}}} \\[2pt]
\multicolumn{7}{c}{\textit{\textbf{Post-Training (hidden\_states alignment whole steps)}}} \\[2pt]
+Corrector(6k) &2.14x&2.390&0.140&0.825&24.017& 57.82\% \\
+Corrector(12k) &2.14x&2.387&0.137&0.827&24.102& 57.76\% \\
+Corrector(18k) &2.15x&2.368&0.140&0.827&24.018& 58.20\% \\
\addlinespace[0.3em]
\multicolumn{7}{c}{\textit{\textbf{Post-Training (output\_velocity alignment whole steps)}}} \\[2pt]
+Corrector(1k) &2.13x&2.392&0.140&0.827&23.833& 57.71\% \\
+Corrector(2k) &2.12x&2.403&0.143&0.825&23.654& 57.46\% \\
+Corrector(3k) &2.13x&2.394&0.148&0.820&23.452& 57.67\% \\
\addlinespace[0.3em]
\multicolumn{7}{c}{\textit{\textbf{Post-Training (output\_velocity alignment cache steps)}}} \\[2pt]
+Corrector(3k) &2.14x&2.379&0.135&0.839&24.325& 58.23\% \\
+Corrector(5k) &2.15x&2.370&0.137&0.839&24.359& 58.08\% \\
+Corrector(10k) &2.14x&2.384&0.140&0.837&24.284& 57.91\% \\
\addlinespace[0.3em]
\hline
\multicolumn{7}{c}{\textit{\textbf{Sequential Training: Policy (40k) + Corrector (12k)}}} \\[2pt]
Sequential &3.05x&1.673&0.254&0.730&21.042& 73.17\% \\
Sequential &2.75x&1.851&0.212&0.768&22.084& 69.36\% \\
Sequential &2.56x&1.987&0.184&0.793&22.678& 66.48\% \\
Sequential &2.14x&2.388&0.135&0.842&24.499& 57.66\% \\
Sequential &1.92x&2.654&0.109&0.867&25.486& 51.95\% \\
Sequential &1.75x&2.916&0.089&0.886&26.643& 46.42\% \\
Sequential &1.61x&3.161&0.076&0.901&27.531& 41.15\% \\
\addlinespace[0.3em]
\multicolumn{7}{c}{\textit{\textbf{OnlineCache with BLO (Outer+Inner epochs)}}} \\[2pt]
BLO (36k+9k) &2.82x&1.806&0.226&0.758&21.802&70.54\%\\
BLO (36k+9k) &2.62x&1.944&0.197&0.784&22.558&67.64\%\\
BLO (36k+9k) &2.55x&1.997&0.185&0.793&22.792&66.23\%\\
BLO (32k+8k) &2.55x&1.999&0.189&0.790&22.758&66.38\%\\
BLO (28k+7k) &2.96x&1.725&0.245&0.739&21.442&72.22\%\\
BLO (28k+7k) &2.83x&1.780&0.225&0.758&21.917&70.56\%\\
BLO (28k+7k) &2.54x&2.011&0.185&0.794&22.849&65.98\%\\
BLO (28k+7k) &2.23x&2.283&0.147&0.830&24.015&60.20\%\\
BLO (28k+7k) &2.07x&2.463&0.125&0.850&24.768&56.32\%\\
BLO (28k+7k) &1.76x&2.898&0.091&0.881&26.182&46.72\%\\
BLO (28k+7k) &1.58x&3.228&0.081&0.896&27.104&41.19\%\\
BLO (24k+6k) &2.61x&1.951&0.203&0.779&22.152&67.18\%\\
BLO (20k+5k) &2.58x&1.979&0.199&0.781&22.175&66.68\%\\
\hline
\end{tabular}
}
\end{minipage}
\end{table*}

The left panel of Table~\ref{tab:appendix2_experiment_results} details the performance evolution of the vanilla OnlineCache method on \textsc{FLUX.1-dev} model at various training checkpoints. We observe three distinct phases:
\begin{itemize}[label=\(\triangleright\), leftmargin=*, itemsep=2pt]
    \item \textbf{Underfitting Phase (12k--36k steps):} The model exhibits lower visual quality as the caching policy has not yet fully adapted to the diffusion trajectory.
    \item \textbf{Optimal Range (38k--40k steps):} The model achieves the best speed-quality trade-off.
    \item \textbf{Overfitting Phase (42k+ steps):} Extended training leads to overfitting, where the policy becomes too aggressive or rigid, resulting in a degradation of visual fidelity.
\end{itemize}

The right panel of Table~\ref{tab:appendix2_experiment_results} reports \textbf{supporting quantitative results}. 
Specifically, the upper block "\textbf{\textit{OnlineCache with Corrector Post-Training}}" examines different corrector training targets. The bottom block demonstrates that the proposed bilevel optimization framework consistently outperforms sequential training pipeline. Detailed analysis of both are provided in Section~\ref{appendix2_ablation_on_corrector_training_target}.

\begin{table}[h]
    \centering
    \scriptsize
    \caption{\textbf{Implementation details and experimental configuration.}}
    \label{tab:experiment_configuration}
    \renewcommand{\arraystretch}{1.1}
    \setlength{\tabcolsep}{2pt}
    \begin{tabular}{ll}
        \toprule
        \textbf{Category} & \textbf{Configuration} \\
        \midrule
        \multicolumn{2}{l}{\textbf{\textit{Hardware \& System}}} \\
        GPU / CUDA / Python & NVIDIA A100-SXM4-80GB ($\times$1) / 12.1 / 3.10.12 \\
        \midrule
        \multicolumn{2}{l}{\textbf{\textit{Core Libraries}}} \\
        PyTorch / Diffusers & 2.3.0+cu121 / 0.34.0 \\
        \midrule
        \multicolumn{2}{l}{\textbf{\textit{Training Hyperparameters}}} \\
        FLUX.1-dev & Grad. Accum.: 30;\quad Batch Size: 1;\quad Loss Weights ($\lambda_{q,a,s,e}$, $k_{\text{tol}}$): (2.5, 10.0, 1.0, 0.5), 6;\quad \\
        DiT-XL/2 & Grad. Accum.: 30;\quad Batch Size: 1;\quad Loss Weights ($\lambda_{q,a,s,e}$, $k_{\text{tol}}$): (2.5, 10.0, 1.0, 0.5), 6;\quad \\
        CogVideoX-2b &Grad. Accum.: 25;\quad Batch Size: 1;\quad Loss Weights ($\lambda_{q,a,s,e}$, $k_{\text{tol}}$): (5.0, 35.0, 1.5, 0.1), 6;\quad \\
        \midrule
        \multicolumn{2}{l}{\textbf{\textit{Baseline Evaluation Settings}}} \\
        FLUX.1-dev & Steps: 30;\quad CFG: 3.5;\quad Resolution: 512×512;\quad Scheduler: \textit{FlowMatchEulerDiscreteScheduler} \\
        DiT-XL/2 & Steps: 30;\quad CFG: 1.7;\quad Resolution: 256×256;\quad Scheduler: \textit{DDIMScheduler} \\
        CogVideoX-2b & Steps: 36;\quad CFG: 6.0;\quad Resolution: 480×720;\quad Scheduler: \textit{CogVideoXDDIMScheduler} \\
        \bottomrule
    \end{tabular}
\end{table}
For the detailed experimental configuration, please check the Table~\ref{tab:experiment_configuration}.

\subsection{Training Dynamics}
\label{appendix2_training_dynamics}
\begin{figure*}[h]
    \centering
    \begin{minipage}[b]{0.66\textwidth}
        \begin{subfigure}[b]{\linewidth}
            \centering
            \includegraphics[width=\linewidth]{fig/ablation_flux_cma_plot.pdf}
            \caption{\textsc{FLUX.1-dev}: Outer Loop (Policy Training)}
            \label{fig:flux_outer}
        \end{subfigure}
        \vspace{2pt}
        \begin{subfigure}[b]{\linewidth}
            \centering
            \includegraphics[width=\linewidth]{fig/alternative_corr_raw_loss_diff.pdf}
            \caption{\textsc{FLUX.1-dev}: Inner Loop (Corrector Training)}
            \label{fig:flux_inner}
        \end{subfigure}
    \end{minipage}
    \hfill
    \begin{minipage}[b]{0.33\textwidth}
        \begin{subfigure}[b]{\linewidth}
            \centering
            \includegraphics[width=\linewidth]{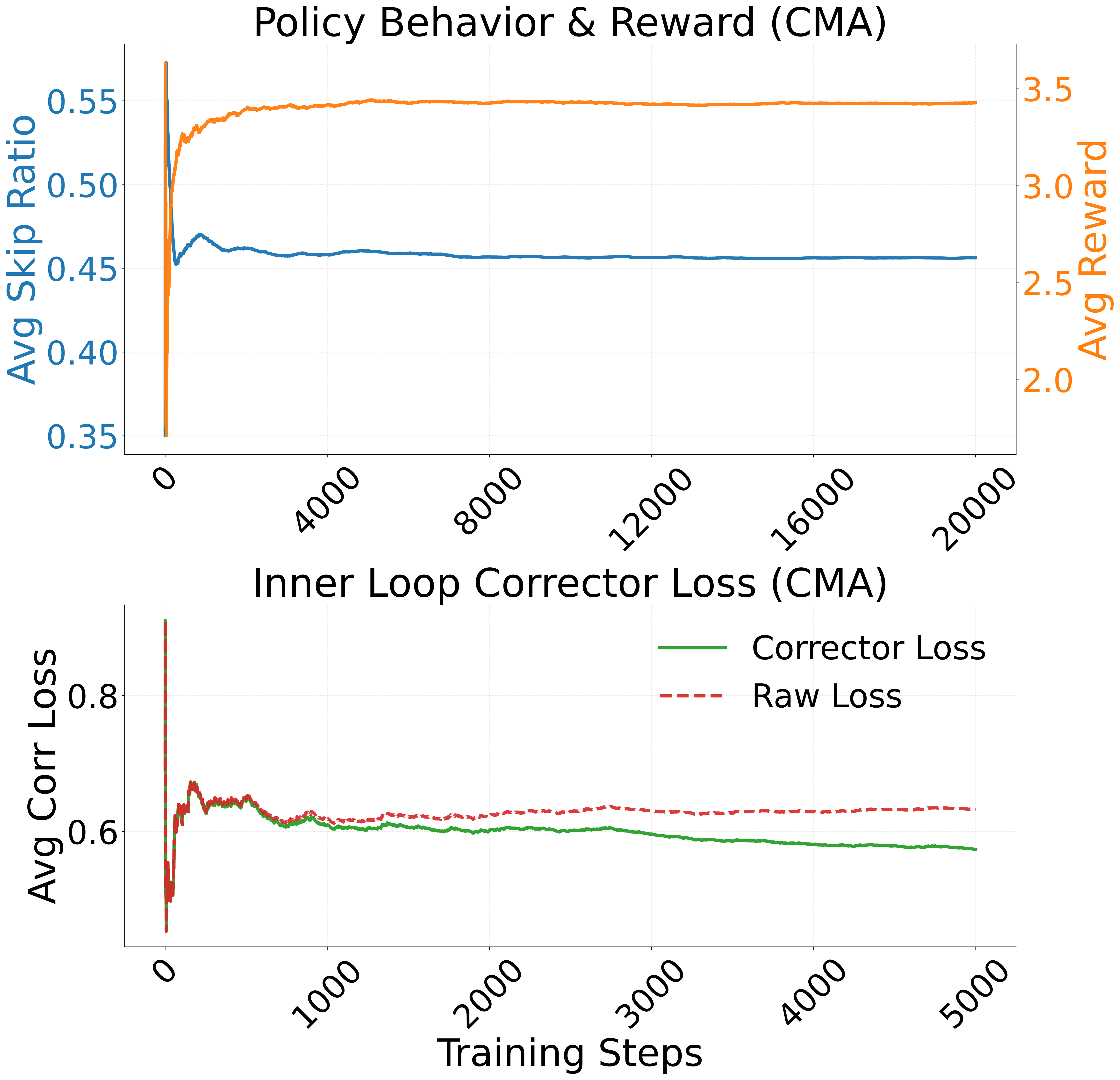}
            \caption{\textsc{DiT-XL/2} Dynamics}
            \label{fig:dit_metrics}
        \end{subfigure}
        \vspace{2pt} 
        \begin{subfigure}[b]{\linewidth}
            \centering
            \includegraphics[width=\linewidth]{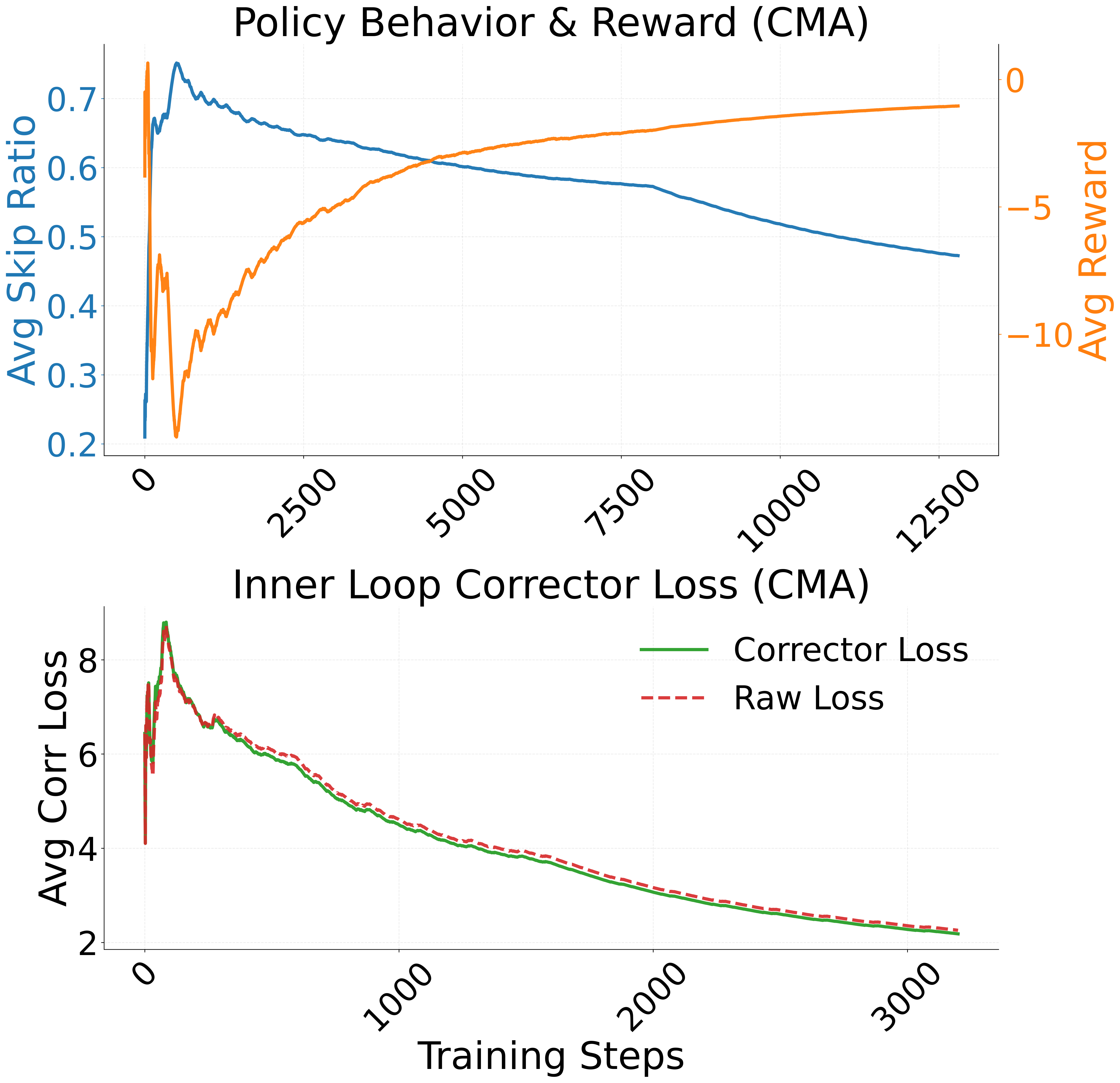}
            \caption{\textsc{CogVideoX-2b} Dynamics}
            \label{fig:cogvideo_metrics}
        \end{subfigure}
    \end{minipage}
    \caption{\textbf{Comprehensive training dynamics of OnlineCache across models and modalities.} 
    (\textbf{a}--\textbf{b}) Drill-down analysis of the bilevel optimization on \textsc{FLUX.1-dev}, showing the evolution of the reward-weighted cache ratio (outer loop) and the rectification of cache-induced errors (inner loop). 
    (\textbf{c}--\textbf{d}) Stable convergence of the policy and corrector on \textsc{DiT-XL/2} (image) and \textsc{CogVideoX-2b} (video). All curves demonstrate that our BLO framework facilitates robust joint optimization.}
    \label{fig:global_training_stability}
\end{figure*}

To examine the convergence behavior of OnlineCache, we analyze the training dynamics of three models, with the corresponding curves visualized in Figure~\ref{fig:global_training_stability}. As shown in~\ref{fig:flux_outer}, the policy exhibits stable convergence behavior during training, with the cache ratio and reward curves gradually stabilizing over iterations, indicating reliable optimization dynamics. 
Figure~\ref{fig:flux_inner} illustrates the inner-loop training dynamics. The error corrector consistently enhances inference performance across training. In particular, the corrected loss (green curve) remains steadily lower than the raw loss (red curve) throughout the entire training process, yielding an average \textbf{error reduction of 11.97\%}. 

We also employ the BLO variant of OnlineCache to accelerate inference for both \textsc{DiT-XL/2}~\citep{DBLP:conf/iccv/PeeblesX23} and \textsc{CogVideoX-2b}~\citep{DBLP:conf/iclr/YangTZ00XYHZFYZ25}. For \textsc{DiT-XL/2}, we utilize the outer-level policy network from the 20k iteration checkpoint and the inner-level error corrector from the 5k iteration checkpoint to conduct the main paper experiments. Regarding \textsc{CogVideoX-2b}, the policy and corrector are selected at 12k (12,800) and 3k (3,200) iterations.

As illustrated in Figure \ref{fig:dit_metrics}, the training process for both models exhibits remarkable stability. The reward CMA (orange) consistently ascends, while the inner-level correction loss (green) shows a steady decline, demonstrating robust convergence across different architectures. 
Specifically, in the bottom panel of Figure \ref{fig:cogvideo_metrics}, the correction loss (green) consistently remains below the raw loss (red), underscoring the effectiveness of the corrector. While the gap in the right plot appears visually narrower compared to the DiT results, this is primarily an artifact of the y-axis scaling: the initial training fluctuations in CogVideoX result in a larger range, whereas the DiT plot focuses on a much finer scale. Despite these visualization differences, both models confirm that the BLO framework achieves reliable and stable convergence.

\vspace{1\baselineskip}
\section{Limitations}
\label{appendix3_limitations}
While OnlineCache provides a robust, instance-aware framework for dynamic caching in diffusion inference, we identify the following limitations that outline promising directions for future research:
\begin{itemize}[label=\(\triangleright\), leftmargin=*, itemsep=2pt]
    \item \textbf{Training Overhead Requirements:} Unlike purely heuristic caching strategies that operate in a strict training-free manner, OnlineCache requires an upfront training phase to optimize the policy and corrector. Although this one-time computational cost is empirically manageable as discussed, it still introduces an additional pipeline step before deployment. 
    \item \textbf{Approximations in Bilevel Optimization:} The exact computation of implicit gradients in our Bilevel Optimization (BLO) framework is intractable. Furthermore, unrolling the full diffusion trajectory for high-resolution models incurs a prohibitive memory footprint ($>$400GB). Consequently, we employ a first-order approximation alongside Truncated Backpropagation Through Time. While empirically effective and memory-efficient, this alternating coordinate-descent approximation may theoretically converge to a suboptimal local equilibrium compared to an exact unrolled solution.
    \item \textbf{Cross-Architecture Dependency:} As demonstrated in Section~\ref{ex:generalization}, the learned modules exhibit strong zero-shot generalization across varying resolutions, dataset distributions, and closely aligned models (e.g., \textsc{FLUX.1-dev} to \textsc{FLUX.1-schnell}). However, because the policy state vector $s_t$ natively depends on channel-wise statistics ($D$-dimensional) tied to a specific latent space, transferring the framework to a fundamentally different backbone architecture (e.g., adapting a DiT-trained policy to a standard U-Net) inherently necessitates retraining from scratch. 
\end{itemize}

\vspace{1\baselineskip}
\section{Broader Impacts}
\label{appendix4_broader_impacts}
Our work focuses on accelerating the inference of diffusion models. A primary positive societal impact of this research is the significant reduction in computational resources and energy consumption required for high-fidelity generation, which directly contributes to lowering the carbon footprint of deploying large-scale generative AI systems. Furthermore, by improving inference efficiency, our method helps democratize access to state-of-the-art generative models for researchers and end-users with limited compute budgets. 

On the potential negative side, making high-quality image and video generation faster and cheaper could inadvertently facilitate the production of malicious content, such as disinformation or deepfakes. However, our method is fundamentally a generic algorithmic optimization; it does not introduce new capabilities to the base models. The mitigation of such malicious uses relies on existing safety guardrails and ethical deployment strategies implemented at the foundation model level.

\clearpage
\section{Qualitative Visual Results}
\label{appendix5_visual_results}
\begin{figure}[h]
    \centering
    \includegraphics[width=\columnwidth]{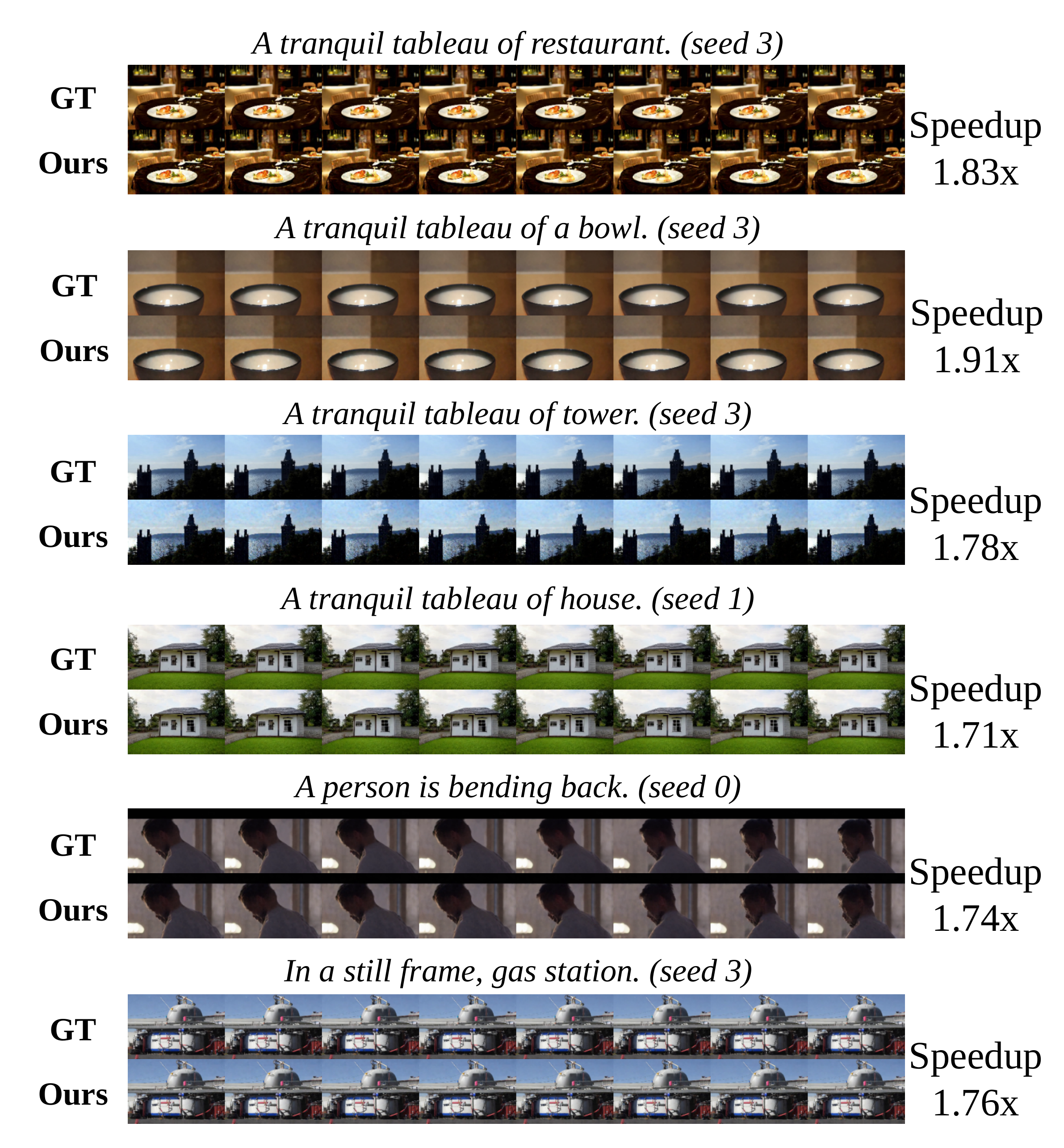}
    \caption{
    Qualitative visualization of generation quality using the \textsc{CogVideoX-2b}~\citep{DBLP:conf/iclr/YangTZ00XYHZFYZ25} model. 
    Our method consistently yields high-fidelity results that are visually indistinguishable from the ground-truth (GT) samples, even while achieving an average speedup exceeding 1.75$\times$.
    }
    \label{fig:visual_cogvideo}
\end{figure}

\clearpage
\begin{figure}[h]
    \centering
    \includegraphics[width=0.95\columnwidth]{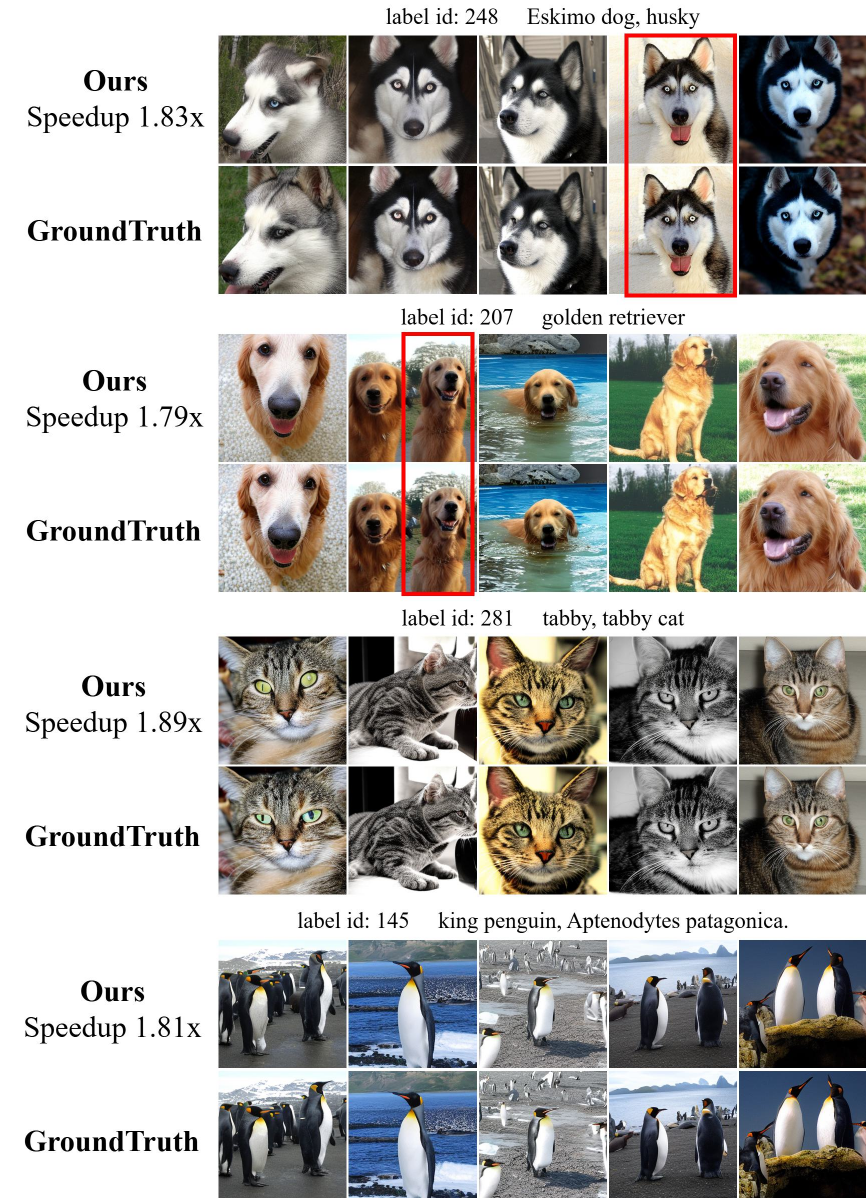}
    \caption{
    Qualitative visualization of generation quality using the \textsc{DiT-XL/2}~\citep{DBLP:conf/iccv/PeeblesX23} model.
    We compare OnlineCache with the non-acceleration baseline under identical sampling settings: DPM-Solver++~\citep{DBLP:journals/ijautcomp/LuZBCLZ25} scheduler with 50 denoising steps, CFG scale of 1.5. Evaluations are conducted on 4 semantic labels. When evaluating all 1k classes with 50k samples using FID, our method achieves a 1.88× acceleration while still outperforming the baseline (quantitative results are reported in the main paper). \textbf{Notably}, our accelerated generation even surpasses the ground-truth (GT) images in certain cases. As highlighted by the red boxes in the first row, the GT image of the husky exhibits noticeable bodily structural defects, whereas our method produces a more complete and coherent structure. For the Golden Retriever in the second row, our model generates fur textures that appear smoother and more natural than GT. 
    }
    \label{fig:visual_dit}
\end{figure}

\clearpage
\begin{figure}[h]
    \centering
    \includegraphics[width=0.95\columnwidth]{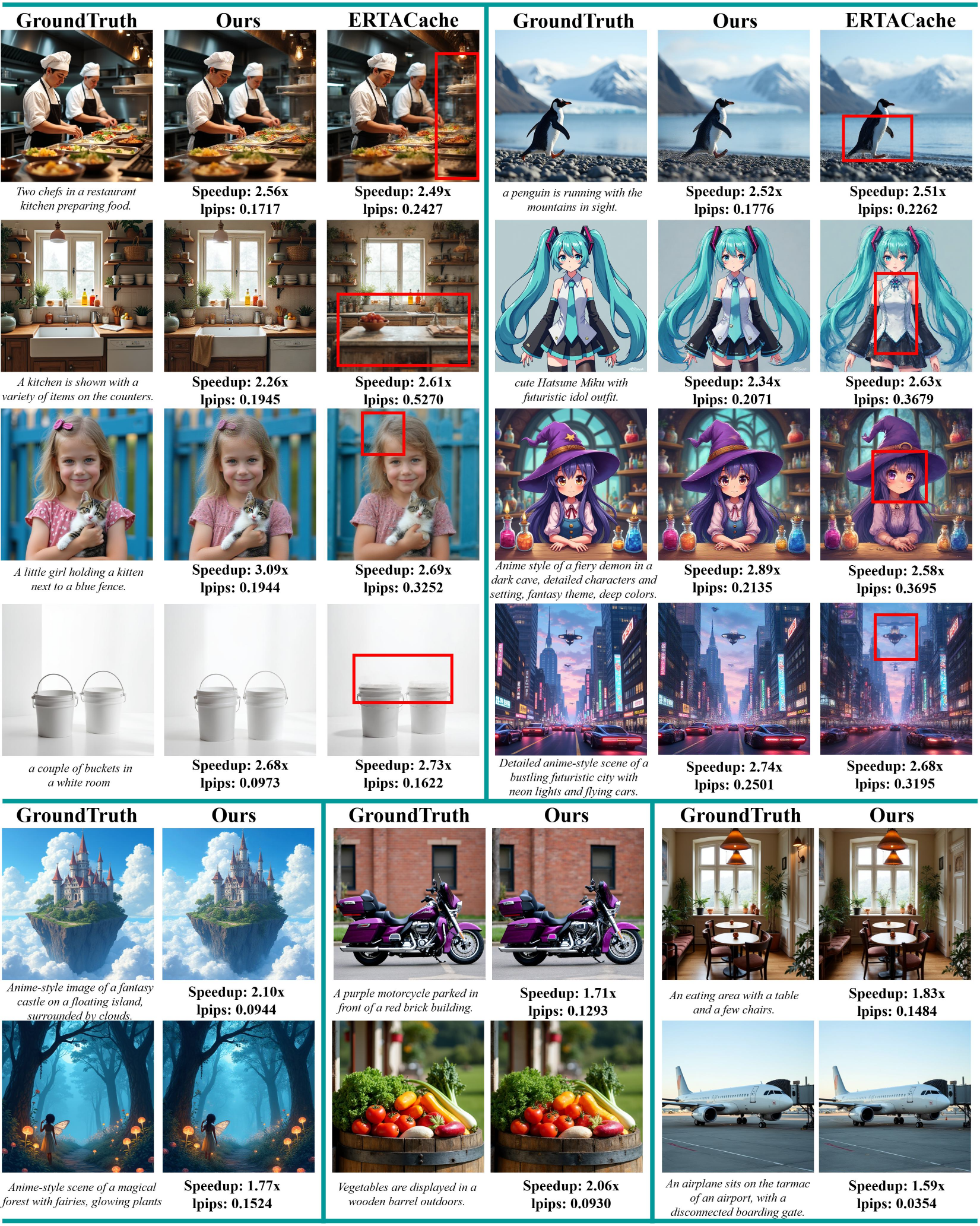}
    \caption{
    Qualitative visualization of generation quality using the \textsc{FLUX.1-dev}~\citep{DBLP:journals/corr/abs-2506-15742} model.
    The upper portion of the figure presents a visual comparison between OnlineCache and ERTACache~\citep{DBLP:journals/corr/abs-2508-21091}, demonstrating that our approach achieves superior visual quality while maintaining effective acceleration. 
    The lower portion further provides additional qualitative results, comparing our generated samples with the corresponding ground-truth images, highlighting the high-fidelity generation of our method. 
    }
    \label{fig:visual_flux}
\end{figure}

\end{document}